%% file: main.tex
\documentclass[lettersize,journal]{IEEEtran}
\usepackage{amsmath,amsfonts}

\usepackage{array}
\usepackage[caption=false,font=normalsize,labelfont=sf,textfont=sf]{subfig}
\usepackage{textcomp}
\usepackage{stfloats}
\usepackage{url}
\usepackage{verbatim}
\usepackage{graphicx}
\usepackage{multirow} 
\usepackage{duckuments}
\usepackage{booktabs}
\usepackage{makecell} 

\usepackage{xcolor}
\usepackage{algorithmic}
\usepackage{algorithm}
\usepackage{cite}
\usepackage{bm}

\definecolor{topicblue}{RGB}{52, 144, 220}
\definecolor{deepred}{RGB}{180, 0, 0}
\definecolor{orange}{RGB}{255, 165, 0}

\usepackage[colorlinks=true,
            linkcolor=orange,   
            urlcolor=magenta,       
            citecolor=topicblue]    
            {hyperref}

\begin{document}

\title{Breaking the Vicious Cycle: Coherent 3D Gaussian Splatting from Sparse and Motion-Blurred Views}
\author{
        Zhankuo Xu,
        Chaoran Feng,
        Yingtao Li,
        Jianbin Zhao, 
        Jiashu Yang,
        Wangbo Yu,
        \\
        Li Yuan$^\dagger$,
        and
        Yonghong Tian$^\dagger$,~\IEEEmembership{Fellow,~IEEE}
        
\IEEEcompsocitemizethanks{
    \IEEEcompsocthanksitem Z. Xu and C. Feng contributed equally to this work.
    \IEEEcompsocthanksitem C. Feng, W. Yu, L. Yuan, and Y. Tian are with Peking University. Z. Xu, J. Zhao, J. Yang, and Y. Li are research interns at the School of Electronics and Computer Engineering, Peking University.
    \IEEEcompsocthanksitem L. Yuan and Y. Tian are the corresponding authors ($\dagger$). E-mail: \{yuanli-ece@pku.edu.cn, yhtian@pku.edu.cn\}.
}
}

\markboth{Journal of \LaTeX\ Class Files,~Vol.~14, No.~8, August~2021}%
{Shell \MakeLowercase{\textit{et al.}}: A Sample Article Using IEEEtran.cls for IEEE Journals}

\maketitle
\input{Sections/0_abstract}
\input{Sections/1_introduction}
\input{Sections/2_related_work}

\input{Sections/2_5_preliminary}
\input{Sections/3_methods}
\input{Sections/4_exp}
\input{Sections/5_limitation_and_conclusion}

\bibliographystyle{IEEEtran}
\bibliography{tscvt}

\clearpage

\input{Sections/X_supplementary}

\end{document}

%% file: Sections/0_abstract.tex
\begin{abstract}
3D Gaussian Splatting (3DGS) has emerged as a state-of-the-art method for novel view synthesis. 
However, its performance heavily relies on dense, high-quality input imagery, an assumption that is often violated in real-world applications, where data is typically sparse and motion-blurred.
These two issues create a vicious cycle: sparse views ignore the multi-view constraints necessary to resolve motion blur, while motion blur erases high-frequency details crucial for aligning the limited views. 
Thus, reconstruction often fails catastrophically, with fragmented views and a low-frequency bias.
To break this cycle, we introduce CoherentGS, a novel framework for high-fidelity 3D reconstruction from sparse and blurry images. 
Our key insight is to address these compound degradations using a dual-prior strategy. 
Specifically, we combine two pre-trained generative models: a specialized deblurring network for restoring sharp details and providing photometric guidance, and a diffusion model that offers geometric priors to fill in unobserved regions of the scene.
This dual-prior strategy is supported by several key techniques, including a consistency-guided camera exploration module that adaptively guides the generative process, and a depth regularization loss that ensures geometric plausibility.
We evaluate CoherentGS through both quantitative and qualitative experiments on synthetic and real-world scenes, using as few as 3, 6, and 9 input views. Our results demonstrate that CoherentGS significantly outperforms existing methods, setting a new state-of-the-art for this challenging task.
The code and video demos are available at \href{https://potatobigroom.github.io/CoherentGS/}{https://potatobigroom.github.io/CoherentGS/}.
\end{abstract}


%% file: Sections/1_introduction.tex
\section{Introduction}

Recently, 3D Gaussian Splatting (3DGS)~\cite{kerbl3Dgaussians} has emerged as an efficient and expressive scene representation, substantially advancing novel view synthesis (NVS). 
With dense multi-view coverage and sharp observations, 3DGS achieves high-fidelity reconstructions and has been rapidly adopted in a variety of applications~\cite{deblur-nerf,oh2024deblurgs,li2025beyond,chen2024deblurgs,lee2024deblurring_3d_gs,scaffold-gs,tang2024cycle3d,li2025decoupled,wang2024cove,fan2024instantsplat,feng2024dit4edit,zhang2023repaint123,bao2025-3dgs-survey-tcsvt25,zhao2025tune}. 
However, such successes rely on input conditions that rarely hold in practice. In real-world scenarios, particularly in quick and unconstrained handheld/robot-mounted capture, image sequences are often sparse in viewpoints and degraded by motion blur.
Under these conditions, the assumptions underpinning 3DGS break down: sparse views lead to fragmented local representations that fail to consolidate into a coherent global scene~\cite{cheng2025reggs,liu2024georgs-tcsvt24,evagaussians,linhqgs}, while motion blur suppresses high-frequency details that are crucial for reliable geometry and texture recovery~\cite{zhang2024fregs,tang2025neuralgs}. 
When combined, these factors severely underconstrain optimization, making blur disentanglement particularly ill-posed and leading to severe degradation in reconstruction quality.

\begin{figure}[!t]
    \centering
    \includegraphics[width=\linewidth]{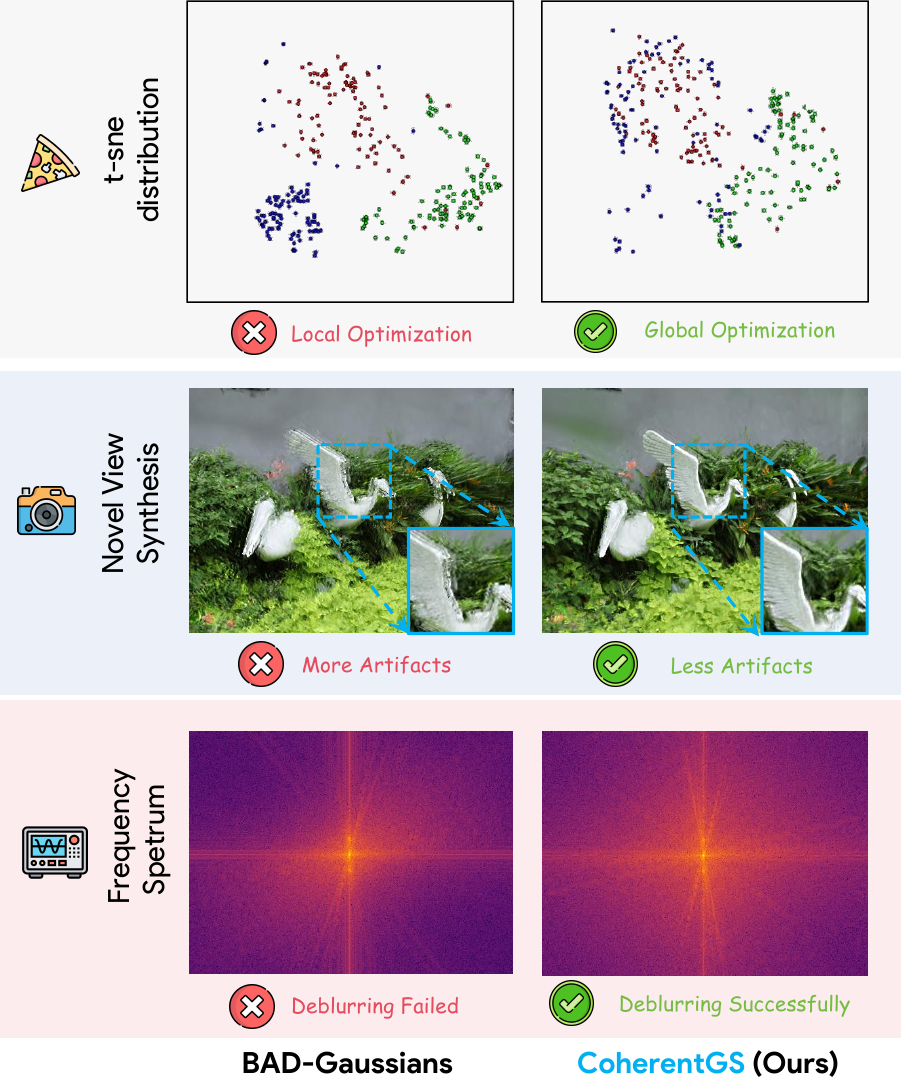}
    \caption{    
    \textbf{Visualizing view fragmentation.} 
    We compare NVS and t-SNE distributions using only 3 input views. 
    BAD-Gaussians\cite{zhao2024badgs} degenerates into fragmented local clusters, whereas CoherentGS maintains global geometric coherence. 
    }
    \label{fig:tsne_analysis}
\end{figure}

These challenges have two main consequences for 3D reconstruction: 
\textbf{view fragmentation} and \textbf{low-frequency bias}. 
Under sparse viewpoints, limited overlap weakens cross-view constraints and drives each camera to fit a localized distribution of Gaussian primitives rather than contributing to a coherent global representation; empirically, the learned distributions form fragmented clusters with weak cross-view overlap, leading to geometric drift and visible inconsistencies across views as shown in Fig.~\ref{fig:tsne_analysis}. 
Motion blur suppresses high-frequency cues such as textures and edges; instead of recovering these details, optimization typically reduces reprojection error by inflating Gaussian covariances or blending colors, thereby fitting low-frequency signals at the expense of structural fidelity~\cite{dai2008motion-blur-formation,son2022real-time-video-deblur,nie2020-video-deblurring-interp-tcsvt2020}. 
Fundamentally, fragmentation and low-frequency bias jointly exacerbate the ill-posedness: degenerate view-specific solutions explain blur via over-smoothed parameterizations, and the loss of high-frequency cues weakens geometric alignment across views.
This vicious coupling leaves the problem severely underconstrained, preventing high-quality novel-view synthesis and stable scene-level coherence, consequently photorealistic supervision~\cite{kerbl3Dgaussians} and exposure modeling~\cite{zhao2024badgs,chen2024deblurgs,wang2023badnerf} alone are inadequate in this regime.

To address these challenges, we introduce a novel 3DGS-based reconstruction framework, \textbf{CoherentGS}. 
We posit that in the challenging real-world setting of sparse and blurry inputs, existing paradigms are insufficient.
On one hand, state-of-the-art 3D deblurring methods~\cite{deblurnerf,zhao2024badgs} excel at modeling camera motion but fundamentally rely on dense multi-view geometric constraints to decouple the trajectory from the scene. 
Under sparse views, these constraints vanish, causing the motion estimation to become ill-posed and the deblurring process to fail. 
On the other hand, while generative models can fill in missing content for sparse views, they are not specialized for physically-grounded deblurring and can hallucinate details inconsistent with the underlying motion. 
Therefore, our core insight is that \textit{a robust solution must synergistically integrate two distinct, specialized priors: one for photometric deblurring and another for geometric completion}.

Our CoherentGS framework employs this dual-prior strategy through a systematic and iterative process. 
The foundation of our method is a camera exposure motion model that explicitly simulates the blur formation by integrating multiple sharp 3DGS renderings along an optimized camera trajectory. 
To provide robust deblurring guidance, we distill knowledge from a pre-trained image deblurring model into our 3DGS representation using a perceptual distillation loss, which enforces the recovery of high-frequency details. 

For scene-level geometric completion, we first propose a consistency-guided trajectory planner that identifies under-observed areas. 
A powerful diffusion model is then used to provide a supervisory signal for these novel views via a score distillation-inspired loss, ensuring contextual and geometric coherence. 
Finally, all components are optimized jointly within a unified framework: a composite objective combines a blurry reconstruction loss on real inputs with the two distillation losses from our generative priors, simultaneously refining the 3DGS scene and the camera motion parameters. 
This iterative loop of guided synthesis and joint optimization progressively enhances the scene's completeness and high-frequency detail.
We evaluate our method on the challenging Deblur-NeRF dataset~\cite{deblur-nerf} and proposed outdoor motion-blurred dataset.
The experimental results demonstrate significant improvements over existing approaches and our contributions are summarized as follows:

\begin{itemize}
    \item We introduce the first 3D Gaussian Splatting framework to systematically address the compound challenge of high-quality reconstruction from inputs that are simultaneously sparse and degraded by severe motion blur.
    
    \item We propose a novel dual-prior guidance strategy that integrates a generative diffusion model to resolve sparse-view ambiguities and a dedicated deblurring model to restore high-frequency details corrupted by motion.
    
    \item We develop a consistency-guided camera exploration for holistic and unseen view modeling, an exposure photometric bundle adjustment tailored for motion-blurred images, and a training scheme that jointly optimizes 3DGS with generated sequences.
\end{itemize}

%% file: Sections/2_related_work.tex
\section{related work}

\subsection{3D Representation from Degraded Images}
Neural Radiance Fields (NeRF)~\cite{nerf} and 3D Gaussian Splatting (3DGS)~\cite{kerbl3Dgaussians} have emerged as powerful and novel 3D representations. 
However, their success is predicated on stringent input requirements, demanding both dense multi-view coverage and high-fidelity, blur-free imagery.
In practical applications, these ideal conditions are seldom met, as captured images are often degraded. 
Consequently, reconstructing 3D scenes from such degraded inputs has become a critical and burgeoning area of research.
Among various degradations, motion and defocus blur are particularly common in multi-view captures. Pioneering work in this domain focused on adapting NeRF to handle such artifacts. 
DeblurNeRF~\cite{deblur-nerf} was the first to jointly deblur and reconstruct a scene, modeling the blur kernel as a per-pixel ray transformation. 
Building on this, subsequent methods sought to impose greater physical consistency; for instance, DP-NeRF~\cite{lee2023dp} modeled the blur as a function of 3D rigid body motion, more closely approximating the physical camera capture process. 
Other notable contributions in this area include ExBluRF~\cite{lee2023exblurf}, PDRF~\cite{peng2023pdrf}, and their variants~\cite{wang2023badnerf,feng2025ae-nerf}. 
With the advent of 3DGS, which offers efficient rendering and an explicit representation, the focus has shifted towards deblurring this new paradigm. 
Methods such as BAD-Gaussians~\cite{zhao2024badgs}, DeblurGS~\cite{chen2024deblur-gs-acmmm24}, Deblur3DGS~\cite{lee2024deblurring3dgs-eccv24}, and others~\cite{lu2025bardgs,bui2025mobgs,feng2025e4dgs,wu2024deblur4dgs} have all proposed effective solutions, typically by explicitly modeling the camera's motion trajectory during the exposure time to deconvolve the blur.

While previous methods have shown significant success, they often rely on the assumption that a relatively dense set of blurry images is available. 
However, this setup does not fully reflect the challenges encountered in real-world scenarios, where data can be both severely degraded and extremely limited. 
A more realistic and challenging situation arises when only a small number of images are available for a given scene, each suffering from motion blur~\cite{sparse-derf-tpami2025}.
We address this gap by tackling the combined problem of 3D reconstruction from inputs that are both sparse and blurry. 
We propose a framework that integrates deblurring techniques with strategies designed for sparse-view settings, thus improving the applicability of 3D reconstruction in real-world conditions.

\subsection{Generative Priors for Novel View Synthesis.}
Leveraging powerful and pre-trained generative models to overcome the limitations of 3D reconstruction from sparse or degraded imagery is a growing research area. 
This trend has largely branched into two specialized sub-domains targeting different types of degradation. 
The first and most prominent sub-domain addresses sparsity, where diffusion models are employed to rectify geometric and photometric deficiencies. 
Recent works including Difix3D+~\cite{wu2025difix3d+}, GSFixer~\cite{yin2025gsfixer}, 3D-Enhancer~\cite{luo2025-3denhancer} and~\cite{zhong2025taming-Video-Diffusion-Prior-with-Scene-Grounding-Guidance,yu2024viewcrafter,liu2024reconx,wu2024reconfusion,wu2025genfusion,paliwal2025ri3d,wei2025gsfix3d,weber2024nerfiller,li2025sparsegs-w,yu2025trajectorycrafter} typically operate by rendering a flawed view, enhancing it with the diffusion prior, and distilling the refined details back into the 3D representation. 
On the other hand, some methods such as Sparse-DeRF~\cite{sparse-derf-tpami2025} to tackle image blur, integrate specialized restoration networks to explicitly reverse the degradation process. 

Despite their individual successes, these two paradigms have remained isolated.
This separation presents a critical challenge: directly applying sparse-view generative methods to blurry inputs leads to hallucinated blur where the model mistakes artifacts for intrinsic texture, while applying deblurring networks to sparse data results in severe overfitting.
Drawing inspiration from 2D knowledge distillation~\cite{zhang2025knowledge-distill-icassp25}, our work is the first to bridge this gap by synergistically integrating distinct types of generative priors within a single, coherent optimization framework.

%% file: Sections/2_5_preliminary.tex
\section{Preliminary}
\label{sec:preliminary}

\subsection{3D Gaussian Splatting}
3DGS~\cite{kerbl3Dgaussians} explicitly represents the scene using a collection of sparse 3D Gaussians. Each primitive is parameterized by a 3D covariance matrix $\mathbf{\Sigma}\in\mathbb{R}^{3\times3}$, a mean position $\boldsymbol{\mu}\in\mathbb{R}^3$, an opacity $o \in [0, 1]$, and spherical harmonics (SH) coefficients for view dependent color. 
The spatial influence of a Gaussian is defined as:
\begin{equation}
    \mathcal{G}(\mathbf{x}) = \exp\left(-\frac{1}{2}(\mathbf{x}-\boldsymbol{\mu})^{\top}\mathbf{\Sigma}^{-1}(\mathbf{x}-\boldsymbol{\mu})\right),
\end{equation}
where $\mathbf{\Sigma}$ is factorized into a scaling matrix $\mathbf{S}\in\mathbb{R}^{3}$ and a rotation matrix $\mathbf{R} \in \mathrm{SO}(3)$ to ensure positive semi definiteness, formulated as $\mathbf{\Sigma}=\mathbf{R}\mathbf{S}\mathbf{S}^{\top}\mathbf{R}^{\top}$. 
To render the scene from a specific viewpoint, the 3D covariance is projected into 2D image space as $\mathbf{\Sigma}^{\prime}=\mathbf{J} \mathbf{W}\mathbf{\Sigma} \mathbf{W}^{\top} \mathbf{J}^{\top}$, where $\mathbf{J}$ is the Jacobian of the affine projective approximation and $\mathbf{W}$ is the viewing transformation matrix. 
The final pixel color $C$ is computed by alpha blending $N$ ordered primitives overlapping the pixel:
\begin{equation}
    C = \sum_{i=1}^N T_i \alpha_i \mathbf{c}_i, \quad \text{with} \quad T_i = \prod_{j=1}^{i-1} (1 - \alpha_j),
\end{equation}
where $\mathbf{c}_i$ is the color derived from SH coefficients, and $\alpha_i$ denotes the final alpha contribution computed by multiplying the learned opacity $o_i$ with the 2D Gaussian evaluation.

\subsection{Image Blur Formation}
\label{sec:pre:image_blur_formation}
The physical image formation process involves accumulating photons over a finite exposure period. Mathematically, this is modeled as the temporal integration of instantaneous sharp irradiance. 
Specifically, the motion blurred image $B(\mathbf{u})$ at pixel coordinate $\mathbf{u}$ is formulated as:
\begin{equation}
\label{eq_continuous_blur_im_formation}
	B(\mathbf{u}) = \frac{1}{\tau} \int_{0}^{\tau} I(t, \mathbf{u}) \, \mathrm{d}t,
\end{equation}
where $\tau$ denotes the camera exposure duration, and $I(t, \mathbf{u})$ represents the instantaneous latent sharp image at time $t \in [0, \tau]$. The normalization factor $1/\tau$ ensures energy conservation during the integration window.
For computational feasibility, we approximate this continuous integral using a discrete summation of $n$ virtual sharp samples:
\begin{equation}\label{eq_blur_im_formation}
	B(\mathbf{u}) \approx \frac{1}{n} \sum_{k=0}^{n-1} I_k(\mathbf{u}),
\end{equation}
where $I_k(\mathbf{u})$ denotes the $k^{\text{th}}$ virtual sharp instance sampled along the camera trajectory within the exposure interval.

\subsection{Diffusion Model}
Diffusion Models (DMs) are generative frameworks designed to approximate a data distribution $p_{\text{data}}(\mathbf{x})$ by reversing a gradual noise addition process. 
In the forward pass, an input sample $\mathbf{x}_0$ is progressively perturbed across discrete timesteps $t \in [0, T]$ by injecting Gaussian noise. This yields a noisy latent $\mathbf{x}_{t} = \alpha_{t}\mathbf{x}_0 + \sigma_{t}\boldsymbol{\epsilon}$, where $\boldsymbol{\epsilon} \sim \mathcal{N}(\mathbf{0}, \mathbf{I})$ is standard normal noise, and $\alpha_t, \sigma_t$ define the noise schedule. 
The generative reverse process utilizes a neural network $\mathbf{F}_{\boldsymbol{\theta}}$, termed the denoiser, to estimate the added noise. The parameters $\boldsymbol{\theta}$ are optimized via a denoising score matching objective:
\begin{equation}
\mathcal{L}_{\text{DM}} = \mathbb{E}_{\mathbf{x}_0, t, \boldsymbol{\epsilon}, \mathbf{c}} \left[ \|\boldsymbol{\epsilon} - \mathbf{F}_{\boldsymbol{\theta}}(\mathbf{x}_{t}; t, \mathbf{c})\|_2^2 \right].
\end{equation}
Here, the network $\mathbf{F}_{\boldsymbol{\theta}}$ takes the noisy latent $\mathbf{x}_t$, the timestep $t$, and an optional condition vector $\mathbf{c}$ (e.g., text embeddings or reference images) as input to reconstruct the signal. We employ this conditional generation capability to guide geometric completion in unobserved regions.

%% file: Sections/3_methods.tex
\section{Method}
\begin{figure*}[!t]
    \centering
        \includegraphics[width=1.0\textwidth]{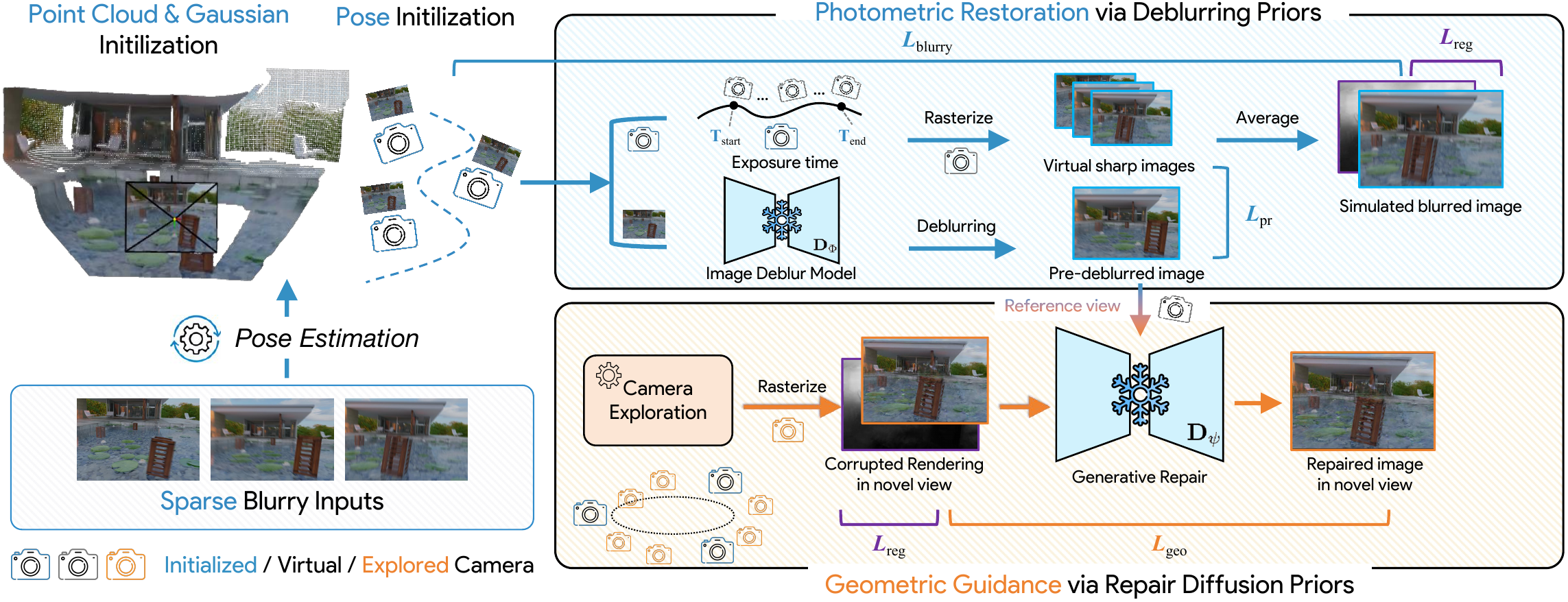}
    \centering
    \caption{
    \textbf{Overview of CoherentGS.} 
    Our framework synergizes two generative priors to resolve sparse view ambiguity and motion blur. 
    (Left) We initialize the Gaussian primitives using poses estimated by COLMAP~\cite{colmap}. 
    (Top) Photometric Restoration via Deblurring Priors: This branch models the physical exposure trajectory to supervise blurry synthesis via $\mathcal{L}_{\text{blurry}}$ and distills sharp high frequency details from a pretrained deblurring network using a perceptual loss $\mathcal{L}_{pr}$. 
    (Bottom) Geometric Guidance via Repair Diffusion Priors: This branch utilizes a diffusion model to repair structural defects in explorative viewpoints. 
    A score distillation loss $\mathcal{L}_{\text{geo}}$ and a depth regularization loss $\mathcal{L}_{\text{reg}}$ are applied to guide the geometry completion and ensure consistency.
    }
    \label{fig:pipeline}
\end{figure*}
\subsection{Problem Formulation and Overall Framework}

Our \textbf{CoherentGS} framework employs a dual prior strategy to address the compound degradation of \textit{view fragmentation} and \textit{low frequency bias} inherent in motion blurred sparse view modeling, as depicted in Fig.~\ref{fig:pipeline}. 
Given only a set of $K$ sparse and blurry images $\{B_i\}_{i=1}^{K}$, we first establish an initial geometric backbone by estimating camera poses $\{\mathbf{T}_i\}_{i=1}^{K}$ by COLMAP~\cite{compactgs_cvpr24} and initializing a 3DGS model $\mathcal{G}_{\theta}$.
The core of our method is an iterative loop designed to progressively enhance this initial reconstruction by synergistically applying two distinct generative priors. 
First, to address motion blur, we perform photometric restoration guided by a deblurring prior (Sec.~\ref{sec:method:camera_trajectory}). 
This module explicitly models the physical exposure trajectory to disentangle camera motion from scene appearance, thereby recovering high frequency details. 
Second, to resolve sparse view ambiguity, we introduce geometric completion guided by a conditional diffusion prior (Sec.~\ref{sec:method:3d_enhancement}). 
This process is steered by a consistency-guided camera exploration strategy (Sec.~\ref{sec:method:camera_trajectory_strategy}), which targets unobserved regions to steer the diffusion model in generating plausible geometric and contextual content. 
Finally, these two guidance mechanisms are integrated within a joint optimization scheme (Sec.~\ref{sec:method:joint_optimization}), where the 3DGS model and camera parameters are updated using a composite loss that fuses supervision from the original blurry inputs with the signals from both generative priors. 
This cyclical generation and reconstruction process is iterated to progressively refine the scene fidelity.
\input{Alg/deblurring_priors}

\subsection{Photometric Restoration via Deblurring Prior}
\label{sec:method:camera_trajectory}

Our approach begins by explicitly modeling the physical process of motion blur as shown in Sec.~\ref{sec:pre:image_blur_formation}. 
The blur in a captured image $B$ is the result of integrating light from a dynamic scene along a camera's trajectory over its exposure time $\tau$. We model this by optimizing a start pose $\mathbf{T}_\mathrm{start}$ and an end pose $\mathbf{T}_\mathrm{end}$ for each capture. The latent camera pose $\mathbf{T}_t$ at any time $t \in [0, \tau]$ is derived through linear interpolation in the $\mathrm{SE}(3)$ space:
\begin{equation} 
\label{eq:exposure_pose}
	\mathbf{T}_t = \mathbf{T}_\mathrm{start} \cdot \exp\left(\frac{t}{\tau} \cdot \log\left(\mathbf{T}_\mathrm{start}^{-1} \cdot \mathbf{T}_\mathrm{end}\right)\right).
\end{equation}
By discretizing the exposure time into $n$ samples, we can render a set of virtual sharp images $\{\hat{I}_i(\mathbf{T}_i, \theta)\}_{i=0}^{n-1}$ from the 3DGS model $\mathcal{G}_\theta$. 
Averaging these sharp renderings allows us to synthesize a blurry image $\hat{B}$:
\begin{equation}
\label{eq:pred_blur_image}
\hat{B}(\theta, \mathbf{T}_\mathrm{start}, \mathbf{T}_\mathrm{end}) = \frac{1}{n} \sum_{i=0}^{n-1} \hat{I}_i(\mathbf{T}_i, \theta).
\end{equation}
This formulation allows us to supervise the 3DGS and the camera motion parameters using a photorealistic loss, $\mathcal{L}_{\text{blurry}}$, which measures the difference between the synthesized blurry image $\hat{B}$ and the real captured image $B$:
\begin{equation}
    \label{eq:photorealistic_loss}
    \mathcal{L}_{\text{blurry}} = \lambda_{1}\mathcal{L}_{1}(\hat{B}, B) + \lambda_{\text{SSIM}}\mathcal{L}_{\text{SSIM}}(\hat{B}, B).
\end{equation}

However, under sparse-view conditions, relying solely on this blurry reconstruction loss is insufficient. 
The joint optimization of deblurring and 3D geometry becomes highly ill-posed, as there are not enough multi-view constraints to robustly decouple the camera motion from the scene's appearance. 
A straightforward alternative would be to pre-deblur the input images and use them for direct pixel-wise supervision. 
This approach is prone to failure, as single-image deblurring is itself an ill-posed problem and applying it independently to multiple views often generates results that are not 3D-consistent, leading to severe geometric artifacts.

Inspired by~\cite{sparse-derf-tpami2025}, we address this by distilling knowledge from a powerful, pre-trained image deblurring model~\cite{mpr} in a robust feature space instead of the pixel space. 
First, we generate a set of target sharp images $\{\bar{I}_i\}_{i=1}^{K}$ by applying a pre-trained deblurring model $\mathbf{D}_{\Phi}$ to blurry inputs $\{B_i\}_{i=1}^{K}$:
\begin{equation}
    \label{eq:deblur_image}
    \bar{I}_i = \mathbf{D}_{\Phi}(B_i).
\end{equation}
 
Instead of enforcing strict pixel alignment with these pseudo-targets which may contain artifacts, we leverage them for high-level semantic guidance. 
We define a perceptual restoration loss $\mathcal{L}_{pr}$, that encourages the features of our rendered sharp image $\hat{I}_i$ to match the features of the target deblurred image $\bar{I}_i$. 
To extract robust features, we employ a shared and pre-trained VGG~\cite{simonyan2014vgg} encoder $\Theta_{\text{vgg}}$:
\begin{equation}
    \label{eq:deblur_priors_loss}
    \mathcal{L}_{pr} = ||\Theta_{\text{vgg}}(\hat{I}_i) - \Theta_{\text{vgg}}(\bar{I}_i)||_2^2.
\end{equation}
For this formulation, the rendered sharp image $\hat{I}_i$ is generated using the camera pose at the temporal midpoint $\mathbf{T}_{t=\tau/2}$, which serves as the canonical sharp state of the frame. 
This distillation mechanism provides strong priors for high-frequency details without enforcing strict    pixel-level correspondence.

\input{Alg/3d_enhancement}

\begin{figure*}[!t]
    \centering
    \includegraphics[width=0.9\textwidth]{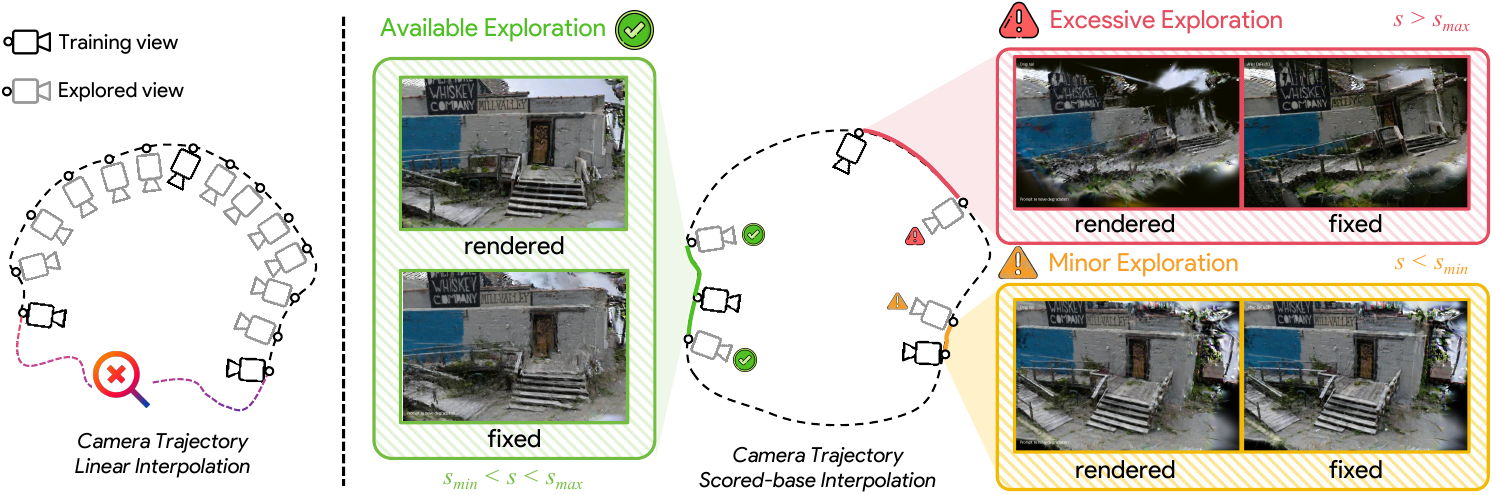}
    \centering
    \caption{\textbf{Comparison of different trajectories. }
    (Left) Linear interpolation between training views provides only limited angular diversity, often leading to ambiguous geometry. 
    (Right) Our scored-based interpolation samples candidate poses along the $\mathrm{SE}(3)$ geodesic and evaluates each with the diffusion-consistency score $\tilde{s}(\mathbf{T})$.
    Valid viewpoints ($s_{\min} \le \tilde{s}(\mathbf{T}) \le s_{\max}$) balance geometric diversity and diffusion stability, while excessive or minor exploration causes instability or redundancy.
    The proposed consistency-guided sampling adaptively expands recoverable viewpoints and enhances novel-view fidelity.
    }
    \label{fig:camera_exploration}
\end{figure*}

\subsection{Geometric Guidance via Diffusion Prior}
\label{sec:method:3d_enhancement}

Recently, leveraging diffusion models to enhance 3DGS from limited inputs has gained significant attention. 
Recent works~\cite{luo2025-3denhancer, wu2025difix3d+, zhong2025taming-Video-Diffusion-Prior-with-Scene-Grounding-Guidance, wu2025genfusion} have shown that these models not only provide additional 3D priors for sparse views but also improve reconstruction quality by introducing high-frequency details. 
In this work, we integrate Difix3D+~\cite{wu2025difix3d+} as our foundational generative prior to rectify structural defects. 
Specifically, it predicts a clean image $\hat{I}^{fix}_i$ by conditioning on a reference view $I_{\text{ref}}$ and the current noisy rendering $\hat{I}_{i}$:
\begin{equation}
\label{eq:difix_enhance}
\hat{I}^{fix}_i = \mathbf{D}_{\psi}(\hat{I}_{i}(\textbf{T}_i,\theta);I_{\text{ref}},t_0),
\end{equation}

Here, we sample a pre-deblurred image $\bar{I}_j$ from Eq.~\ref{eq:deblur_image} as reference image $I_{\text{ref}}$. 
However, directly utilizing these generative models presents a critical challenge in sparse-view scenarios where observations are severely degraded. 
A common approach is to reintroduce the generated images into the 3DGS training pipeline and apply a photorealistic loss for supervision. 
While this naive feedback loop improves sharpness, it introduces a diffusion bias by overemphasizing pixel-wise similarity, leading to over-smoothing and multi-view inconsistencies, especially in sparse regions.

To overcome this limitation, we propose a paradigm shift from using the diffusion model as an offline post-processor to leveraging it as an online supervisory signal via distillation. 

Drawing inspiration from Score Distillation Sampling (SDS)~\cite{poole2022dreamfusion}, our method utilizes the pre-trained 2D diffusion model to provide informative gradients that guide the 3D optimization directly. 
Unlike the original noise-residual formulation in SDS, we adopt a more stable image-residual formulation. 

The core objective is to guide the 3DGS to render images that the diffusion model perceives as structurally valid while maintaining semantic consistency. 
Specifically, we treat the single-step denoised output of the diffusion model as a pseudo-target and minimize its distance to the original 3DGS rendering. For each generated view, we formulate this geometric guidance loss as:
\begin{equation}
    \mathcal{L}_{\text{geo}} = \lambda_{\text{geo}}|| \Theta_{\text{vgg}}(\text{sg}(\hat{I}^{fix}_i)) - \Theta_{\text{vgg}}(\hat{I}_{i}) ||_2^2 
    \label{eq:geo_loss}
\end{equation}

where $\hat{I}_{i}$ is the image rendered from the current state of our 3DGS. 
$\hat{I}^{fix}_i = \mathbf{D}_{\psi}(\hat{I}_{i}; I_{\text{ref}}, t_0)$ is the predicted clean image after a single denoising step by the pre-trained Difix3D+ model, given the rendered image $\hat{I}_i$ and reference image $I_{\text{ref}}$ at a fixed timestep $t_0$. 
The stop-gradient operator, $\text{sg}(\cdot)$, is crucial as it treats the diffusion model's output as a fixed target, ensuring that gradients flow exclusively back to the 3DGS parameters $\theta$. 
Following the official implementation of Difix3D+ configuration, we use a high-noise timestep by setting $t_0=199$.
The detail is outlined in the Alg.~\ref{alg:3d_consistency_distillation_and_guidance}

\subsection{Consistency-guided Camera Exploration}
\label{sec:method:camera_trajectory_strategy}

The efficacy of the diffusion prior (Sec.~\ref{sec:method:3d_enhancement}) critically depends on the selection of viewpoints, where the rendered inputs must be informative enough to guide the geometry while remaining stable enough to prevent generative hallucinations.
The primary challenge lies in exploring novel camera poses that maximize geometric supervision while maintaining structural plausibility.
As illustrated in Fig.~\ref{fig:camera_exploration}, naive linear interpolation between training views limits angular diversity and preserves geometric ambiguities. 
In contrast, large deviations from the image plane frequently result in loss of visual context and significant occlusions.
To balance these conflicting factors, we propose a consistency-guided exploration strategy. 
This consistent-aware mechanism identifies and selects viewpoints that jointly provide high information gains and remain recoverable under the diffusion prior.

\noindent \textbf{Scene Adaptive Consistency Normalization (SACN).}
The absolute scale of the generative consistency score is inherently scene dependent, fluctuating with exposure conditions and blur severity. 
To normalize the metric, we compute a per-scene baseline by averaging diffusion-aided consistency over all training views:
\begin{equation}
\label{eq:coverage_evaluation_train}
\bar{s}=\frac{1}{K}\sum_{i=1}^{K}\Theta_{\mathrm{eval}}\big(\hat{I}^{\mathrm{fix}}(\mathbf{T}_i),\hat{I}(\mathbf{T}_i)\big),
\end{equation}
where $\Theta_{\mathrm{eval}}$ is a pixel-wise evaluator, $\mathbf{T}_i\!\in\!\{\mathbf{T}_i\}_{i=1}^{K}$ are the training poses,  $\hat{I}^{\mathrm{fix}}(T_i)$ is the denoised rendering,  and $\hat{I}(\mathbf{T}_i)$ denotes the pose-specific reference.

\noindent \textbf{Diffusion Reliability Metric.}
For each candidate pose $\mathbf{T}_{new}$, we quantify its diffusion consistency as
\begin{equation}
\label{eq:coverage_evaluation_novel}
s(\mathbf{T}_{new})=\Theta_{\mathrm{eval}}\big(\hat{I}^{\mathrm{fix}}(\mathbf{T}_{new}),\hat{I}(\mathbf{T}_{new})\big),
\end{equation}
Here, a high score suggests the rendering deviates significantly from the diffusion prior, indicating potential artifacts or severe occlusions that the model cannot reliably repair. 
Conversely, an excessively low score implies the view is too similar to the training set or dominated by noise, offering negligible geometric gradients.
To derive a scene agnostic measure, we normalize it by the training baseline $\bar{s}$:
\begin{equation}
\label{eq:score model}
\tilde{s}(\mathbf{T}_{new})=s(\mathbf{T}_{new})-\bar{s}.
\end{equation}
The normalized deviation $\tilde{s}(T)$ serves as a signal for recoverability, indicating how effectively the diffusion prior can refine the rendering without breaking geometric consistency. 

Moderate deviations provide informative geometric cues with high reliability. In contrast, negligible deviations lead to redundant coverage, while excessive deviations risk inducing the unreliable hallucinations.
\input{Alg/camera_exploration}

\noindent \textbf{Band Pass View Selection.}
Candidate poses are generated by densely interpolating on the $\mathrm{SE}(3)$ manifold between adjacent pairs $(\mathbf{T_i}, \mathbf{T_{i+1}})$, supplemented by minor extrapolation. 
This strategy introduces controlled viewpoint perturbations while ensuring the poses remain within vicinity of the observed trajectory.
We retain candidates satisfying the band-pass criterion:
\begin{equation}
s_{\min}\le \tilde{s}(\mathbf{T})\le s_{\max}.
\end{equation}
The selected viewpoints are incorporated into the optimization loop,  adapting the sampling budget toward under-constrained but recoverable regions. 
This targeted exploration progressively enhances novel-view fidelity and stabilizes diffusion-guided supervision with negligible computational overhead.

\subsection{Joint Optimization with Reconstruction and Generation}
\label{sec:method:joint_optimization}
The final stage of our framework involves jointly optimizing the 3D scene representation and the camera motion parameters as shown in Alg.~\ref{alg:joint_optimization}. 
To achieve this, we formulate a composite objective function that integrates supervisory signals derived from both the raw blurry observations and the diffusion rectified novel views. 
This joint optimization allows the model to leverage the ground-truth constraints from the captured data while simultaneously benefiting from the strong generative prior to deblur and complete unseen regions.

To mitigate the inherent geometric ambiguity in sparse view reconstruction, we incorporate a depth regularization loss $\mathcal{L}_{\text{reg}}$ inspired by RegNeRF~\cite{regnerf}. 
This loss enforces a local smoothness prior on the rendered depth maps, which is critical for regularizing geometry in under-constrained and textureless regions.
It effectively prevents degenerate solutions such as surface fragmentation and floating artifacts:
\begin{equation}
    \mathcal{L}_{\text{reg}} = \sum_{i=0}^{n-1} \left( ||\nabla_x \hat{D}_i||_1 + ||\nabla_y \hat{D}_i||_1 \right)
    \label{eq:reg_loss}
\end{equation}

By integrating these complementary supervisory signals, the total objective function $\mathcal{L}_{\text{total}}$ is formulated as a weighted sum of the reconstruction loss, the perceptual restoration loss, the geometric guidance loss, and the regularization term:
\begin{equation}
    \mathcal{L}_{\text{total}} = \mathcal{L}_{\text{blurry}} + \lambda_{\text{pr}}\mathcal{L}_{\text{pr}} + \mathcal{L}_{\text{geo}} + \lambda_{\text{reg}}\mathcal{L}_{\text{reg}}
    \label{eq:total_loss}
\end{equation}
where $\lambda_{\text{pr}}$, $\lambda_{\text{geo}}$ and $\lambda_{\text{reg}}$ are scalar hyperparameters that balance the contribution of each term.

\input{Alg/generation_and_reconstruction}

%% file: Alg/deblurring_priors.tex
\begin{figure}[t]
\begin{algorithm}[H]
\caption{Photometric Restoration via a Deblurring Prior}
\label{alg:perceptual_restoration}
\begin{algorithmic}[1]
\STATE \textbf{Function} ComputePerceptualLoss($\mathcal{G}_{\theta}$, $B_i$, $\mathbf{T}_{\text{start}}$, $\mathbf{T}_{\text{end}}$)
\STATE \textbf{Input:} Current 3DGS model $\mathcal{G}_{\theta}$, a blurry input image $B_i$, and its corresponding start and end poses.
\STATE \textbf{Given:} Pre-trained deblurring model $\mathbf{D}_{\Phi}$, pre-trained feature extractor $\Theta_{\text{vgg}}$.

% \STATE \color{gray}\# --- Generate the sharp target and extract its features ---
\STATE $\bar{I}_i \leftarrow \mathbf{D}_{\Phi}(B_i)$ \hfill $\triangleright$ Eq.~\eqref{eq:deblur_image}.
\STATE $F_{\text{target}} \leftarrow \Theta_{\text{vgg}}(\bar{I}_i)$ 

% \STATE \color{gray}\# --- Render a sharp view from 3DGS at the mean pose ---
\STATE $\mathbf{T}_{\text{mid}} \leftarrow \text{InterpolatePose}(\mathbf{T}_{\text{start}}, \mathbf{T}_{\text{end}}, 0.5)$ \hfill $\triangleright$ Eq.~\eqref{eq:exposure_pose}.
\STATE $\hat{I}_i \leftarrow \text{Render}(\mathcal{G}_{\theta}, \mathbf{T}_{\text{mid}})$ 
\STATE $F_{\text{render}} \leftarrow \Theta_{\text{vgg}}(\hat{I}_i)$ 

% \STATE \color{gray}\# --- Compute the perceptual loss in feature space ---
\STATE $\mathcal{L}_{pr} \leftarrow ||F_{\text{render}} - F_{\text{target}}||_2^2$ \hfill $\triangleright$ Eq.~\eqref{eq:deblur_priors_loss}.

\STATE \textbf{Return} $\mathcal{L}_{pr}$
\end{algorithmic}
\end{algorithm}
\vspace{-4mm}
\end{figure}

%% file: Alg/3d_enhancement.tex
\begin{figure}[t]
\begin{algorithm}[H]
\caption{Geometric Guidance via Diffusion Prior}
\label{alg:3d_consistency_distillation_and_guidance}
\begin{algorithmic}[1]
\STATE \textbf{Function} 3D-Consistency Distillation
\STATE \textbf{Input:} Current 3DGS model $\mathcal{G}_{\theta}$, camera pose $\mathbf{T}_i$, reference image $I_{\text{ref}}$ from vitual viewpoint.
\STATE \textbf{Given:} Pre-trained diffusion denoiser $\mathbf{D}_{\psi}$, noise schedule $\bar{\alpha}_t$, fixed timestep $t_0$, feature extractor $\Theta_{\text{vgg}}$.

\STATE $\hat{I}_{i} \leftarrow \mathcal{G}_{\theta}(\mathbf{T}_i),\mathbf{T}_i\in \{\mathbf{T}_i\}_{i=0}^{n-1}$  \hfill $\triangleright$ Eq.~\ref{eq:exposure_pose}
\STATE $t \leftarrow t_0$ 
\STATE $\boldsymbol{\epsilon} \sim \mathcal{N}(0, \mathbf{I})$ 
\STATE $\hat{I}_{i,t} \leftarrow \sqrt{\bar{\alpha}_t}\hat{I}_{i} + \sqrt{1 - \bar{\alpha}_t}\boldsymbol{\epsilon}$ 
\STATE $\hat{I}^{\text{clean}}_{i} \leftarrow \mathbf{D}_{\psi}(\hat{I}_{i,t}; I_{\text{ref}}, t)$ 
\STATE $\hat{I}^{\text{fix}}_{i} \leftarrow \text{sg}(\hat{I}^{\text{clean}})$ 
\STATE $\mathcal{L}_{\text{geo}} \leftarrow || \Theta_{\text{vgg}}(\text{sg}(\hat{I}^{fix}_i)) - \Theta_{\text{vgg}}(\hat{I}_{i}) ||_2^2 $ \hfill $\triangleright$Eq.~\eqref{eq:geo_loss}
\STATE \textbf{Return} $\mathcal{L}_{\text{geo}}$
\end{algorithmic}
\end{algorithm}
\vspace{-4mm}
\end{figure}

%% file: Alg/camera_exploration.tex
% \begin{algorithm}[t]
% % \fontsize{8.5pt}{10pt}\selectfont  
% % \setstretch{0.92}                  
% \caption{Consistency-guided Camera Exploration}
% \label{alg:unreliable}
% \begin{algorithmic}[1]
% \STATE \textbf{FUNCTION:} 
%  \STATE \quad score calculator: $S(\cdot)$, \hfill $\triangleright$ Eq.~\eqref{eq:score model}
%  \STATE \quad Pose interpolate: $interp(\cdot)$
% \STATE \textbf{Input:}
% Training poses $\{\mathbf{T_i}\}_{i=1}^K$ using VGGT~\cite{wang2025vggt},reference images $\bar{I}_i$ from pre-trained deblur model \hfill $\triangleright$ Eq.~\eqref{eq:deblur_image},
% \STATE \textbf{Given:}
% Pre-trained diffusion denoiser $\mathbf{D}_{\psi}$.Current 3DGS model $\mathcal{G}_{\theta}$,Unreliable threshold $s_{max},s_{min}$, 
% \STATE Append all training views $\mathbf{T_{i}}$ to buffer $\mathcal{B}_{\text{gen}}$
% \FOR{\textbf{each} $\mathbf{T} \in \mathcal{B}_{\text{gen}}$}
%     \STATE $\mathbf{T_{\text{new}}} \leftarrow interp (\mathbf{T_{i}},\mathbf{T_{i+1}})$
%     \STATE $I_{new} \leftarrow \mathcal{G}_{\theta}(\mathbf{T_{new}})$  {render the new pose}
%     \STATE $I_{fix}\leftarrow \mathbf{D}_{\psi}(I_{new},\bar{I}_i,t_0)$ \hfill $\triangleright$ Eq.~\eqref{eq:difix_enhance}
%     \IF{$s_{\min} \le S(I_{\mathrm{fix}}) \le s_{\max}$}
%     \STATE Add $I_{fix}$ to training set.
%     \STATE Add $\mathbf{T_{new}}$ to  buffer $\mathcal{B}_{\text{gen}}$.
%     \ENDIF
% \ENDFOR
% \end{algorithmic}
% \end{algorithm}

\begin{algorithm}[t]
\caption{Consistency-guided Camera Exploration}
\label{alg:unreliable}
\begin{algorithmic}[1]
\STATE \textbf{Input:} Training poses $\{\mathbf{T_i}\}$, Reference images $\{\bar{I}_i\}$ \hfill $\triangleright$ Eq.~\eqref{eq:deblur_image}
\STATE \textbf{Models:} 3DGS $\mathcal{G}_{\theta}$, Diffusion Denoiser $\mathbf{D}_{\psi}$, Score function $S(\cdot)$
\STATE \textbf{Params:} Thresholds $s_{min}, s_{max}$, Interpolation $interp(\cdot)$
\STATE \textbf{Initialize:} Buffer $\mathcal{B}_{\text{gen}} \leftarrow \{\mathbf{T_i}\}$

\FOR{\textbf{each} adjacent pair $(\mathbf{T}_i, \mathbf{T}_{i+1})$ in $\mathcal{B}_{\text{gen}}$}
    \STATE $\mathbf{T_{\text{new}}} \leftarrow interp(\mathbf{T_{i}}, \mathbf{T_{i+1}})$
    \STATE $I_{new} \leftarrow \mathcal{G}_{\theta}(\mathbf{T_{new}})$ 
    \STATE $I_{fix} \leftarrow \mathbf{D}_{\psi}(I_{new}, \bar{I}_i, t_0)$ \hfill $\triangleright$ Eq.~\eqref{eq:difix_enhance}
    
    \STATE $s_{curr} \leftarrow S(I_{\mathrm{fix}})$ \hfill $\triangleright$ Eq.~\eqref{eq:score model}
    
    \IF{$s_{\min} \le s_{curr} \le s_{\max}$}
        \STATE Add $I_{fix}$ to training set
        \STATE Add $\mathbf{T_{new}}$ to $\mathcal{B}_{\text{gen}}$
    \ENDIF
\ENDFOR
\end{algorithmic}
\end{algorithm}

%% file: Alg/generation_and_reconstruction.tex
\begin{figure}[t]
\begin{algorithm}[H]
\caption{Joint Optimization of \textbf{CoherentGS}}
\label{alg:joint_optimization}
\begin{algorithmic}[1]
\STATE \textbf{Input:} Sparse, blurry images $\{B_i\}_{i=1}^K$.
\STATE \textbf{Given:} Training iterations $N_{\text{iters}}$, generation interval $N_{\text{gen}}$, warmup iterations $N_{\text{warmup}}$.
\STATE \textbf{Initialize:}
\STATE \quad Estimate initial poses $\{P_i\}_{i=1}^K$.
\STATE \quad Initialize 3DGS and motion $\theta$, $\{\mathbf{T}_{\text{start},i}, \mathbf{T}_{\text{end},i}\}_{i=1}^K$.
\STATE \quad Initialize a buffer for generated views $\mathcal{B}_{\text{gen}} = \emptyset$.
\FOR{iter = $1, \ldots, N_{\text{iters}}$}
    \STATE \textcolor{gray}{\textit{// Phase 1: Generative Prior Expansion}}

    \IF{iter \% $N_{\text{gen}}$ == 0 \textbf{and} iter $\geq N_{\text{warmup}}$}
        \STATE $\Phi \leftarrow \text{PlanTrajectory}(\mathcal{G}_{\theta}, \{P_i\})$ \hfill $\triangleright$ Sec.~\ref{sec:method:camera_trajectory_strategy}
        \STATE $I_{\text{ref}} \leftarrow$ Render from a training virtual pose.
        \STATE $\mathcal{S}_{\text{new}} \leftarrow \text{EnhanceViews}(\mathcal{G}_{\theta}, \Phi, I_{\text{ref}}, \mathbf{D}_{\psi})$ \hfill $\triangleright$ Sec.~\ref{sec:method:3d_enhancement}
        \STATE Append new views $\mathcal{S}_{\text{new}}$ to buffer $\mathcal{B}_{\text{gen}}$.
    \ENDIF

    \STATE \textcolor{gray}{\textit{// Phase 2: Reconstruction Optimization}}
    \STATE Sample a blurry image $B_i$ and poses $\{\mathbf{T}_{\text{start},i}, \mathbf{T}_{\text{end},i}\}$.
    \STATE $\hat{B}_i \leftarrow \text{RenderBlurImage}(\mathcal{G}_{\theta}, \mathbf{T}_{\text{start},i}, \mathbf{T}_{\text{end},i})$ \hfill $\triangleright$ Eq.~\eqref{eq:pred_blur_image}
    \STATE $\mathcal{L}_{\text{blurry}} \leftarrow \text{ComputeLoss}(\hat{B}_i, B_i)$ \hfill $\triangleright$ Eq.~\eqref{eq:photorealistic_loss}
    \STATE $\mathcal{L}_{pr} \leftarrow \text{ComputeLoss}(\mathcal{G}_{\theta}, B_i, \mathbf{T}_{\text{start},i}, \mathbf{T}_{\text{end},i})$ \hfill $\triangleright$ Eq.~\eqref{eq:deblur_priors_loss}

    \IF{$\mathcal{B}_{\text{gen}}$ is not empty}
        \STATE Sample a generated view $\{\hat{I}_j, \mathbf{T}_j, \hat{D}_j\}$ from $\mathcal{B}_{\text{gen}}$.
        \STATE $\mathcal{L}_{\text{geo}} \leftarrow \text{ComputeLoss}(\mathcal{G}_{\theta}, \hat{I}_j, I_{\text{ref}})$ \hfill $\triangleright$ Eq.~\eqref{eq:geo_loss}
        \STATE $\mathcal{L}_{\text{reg}} \leftarrow \text{ComputeLoss}(\hat{D}_j)$ \hfill $\triangleright$ Eq.~\eqref{eq:reg_loss}
    \ELSE
        \STATE $\mathcal{L}_{\text{geo}} \leftarrow 0$, $\mathcal{L}_{\text{reg}} \leftarrow 0$
    \ENDIF

    \STATE $\mathcal{L}_{\text{total}} \leftarrow \mathcal{L}_{\text{blurry}} + \lambda_{\text{pr}}\mathcal{L}_{\text{pr}} + \lambda_{\text{geo}}\mathcal{L}_{\text{geo}} + \lambda_{\text{reg}}\mathcal{L}_{\text{reg}}$ 
    \STATE Update $\theta$, $\{\mathbf{T}_{\text{start},i}\}$, and $\{\mathbf{T}_{\text{end},i}\}$.
\ENDFOR
\end{algorithmic}
\end{algorithm}
\end{figure}

%% file: Sections/4_exp.tex
\section{Experiments}
\label{sec:exps}

\subsection{Implementation details}
\label{sec:exps:settings}

\noindent \textbf{Dataset and preprocessing.}
We evaluate our method on the standard Deblur-NeRF benchmark~\cite{deblur-nerf}, which comprises 5 synthetic and 5 real-world scenes characterized by severe motion blur.
To further assess the generalization capability in complex, unconstrained outdoor environments, we establish a benchmark \textsc{DL3DV-Blur} based on the DL3DV-10K dataset~\cite{ling2024dl3dv}, simulating realistic motion blur across 5 diverse scenes.
For experimental settings, we adopt sparse-view configurations with $K \in \{3,6,9\}$ views.
We follow the train/test splitting protocol of BAD-Gaussians~\cite{zhao2024badgs}: for the Deblur-NeRF dataset, we adopt the same train/test split as in their setting, while for the DL3DV-Blur dataset, we hold out every 7th image as the test set of novel views.
More details of proposed dataset are in the \textit{appendix}.

\vspace{+6pt}
\noindent \textbf{Evaluation metrics.}
We utilize several metrics to assess the quality of reconstruction and reenactment. The peak signal-to-noise ratio (PSNR), Structural Similarity (SSIM), and Learned Perceptual Image Patch Similarity (LPIPS)~\cite{zhang2018unreasonable-lpips} are employed for evaluating image synthesis quality in novel view synthesis and deblurring view synthesis. 

\vspace{+6pt}
\noindent \textbf{Training Details.}
Our framework is implemented based on the official implementation of BAD-Gaussians~\cite{zhao2024badgs}, incorporating the pre-trained prior from Difix3D+\cite{wu2025difix3d+}.
We employ the Adam optimizer for all learnable parameters.
The learning rate schedules and densification strategies for 3D Gaussian primitives strictly follow the default configuration in~\cite{kerbl3Dgaussians,zhao2024badgs}.
For the camera trajectory modeling (Eq.~\eqref{eq:exposure_pose}), the learning rates for translation and rotation ($T_{\text{start}}, T_{\text{end}}$) are initialized at $5 \times 10^{-3}$ and exponentially decayed to $5 \times 10^{-5}$.The number of virtual camera poses $n$ in (Eq.~\eqref{eq:pred_blur_image}) is set to 10. Regarding the loss terms, we set the blurry weight to $\lambda_{\text{1}} = 0.8, \lambda_{\text{D-SSIM}} = 0.2$ and depth regularization weight $\lambda_{\text{reg}} = 0.1$. The weights for the deblurring ($\lambda_{\text{pr}}$) and geometric ($\lambda_{\text{geo}}$) priors are set to 0.01.
To adapt to different scene distributions, the confidence thresholds for the score model are calibrated per dataset: we set $\{s_{\max}, s_{\min}\}$ to $\{14.5, 4.5\}$ for the outdoor DL3DV-Blur dataset, and $\{8.5, 2.5\}$ for the Deblur-NeRF dataset.The $\Theta_{\mathrm{eval}}$ in (Eq.~\eqref{eq:coverage_evaluation_train} and Eq.~\eqref{eq:coverage_evaluation_novel}) are defined as $psnr(\cdot)$.
The training initiates with a warm-up phase of 1500 iterations dedicated to the deblurring model.
Subsequently, we incorporate diffusion prior and optimize under the guidance of the camera trajectory model, with the interpolation interval set to 200 steps.
In total, the training spans a total of 7,000 iterations.
All experiments are performed on a NVIDIA A6000 with 48GB memory.
More training details are in the \textit{appendix}.

\input{Tables/table_exp_deblur}
\begin{figure*}[!ht]
    \centering
        \includegraphics[width=0.95\textwidth]{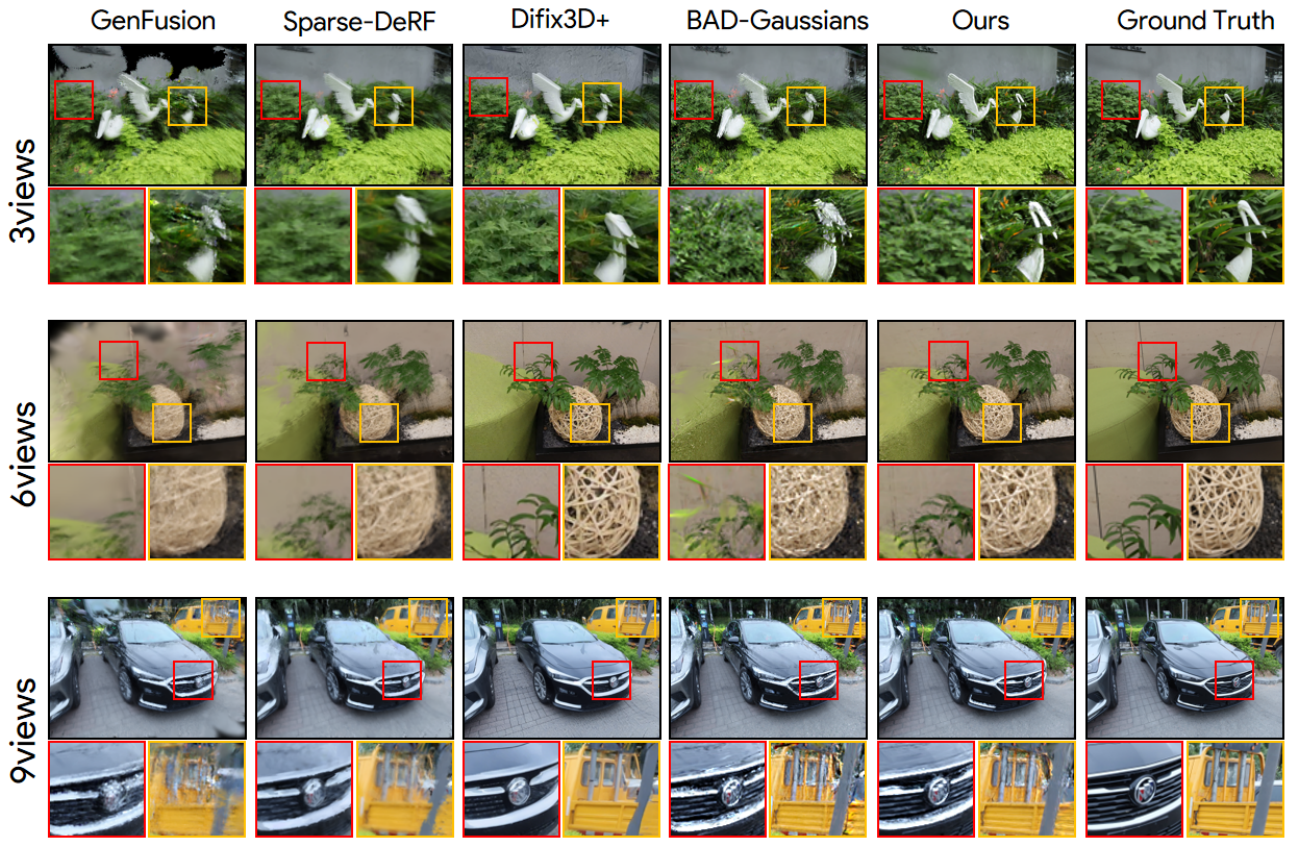}
    \centering
    \caption{
         \textbf{Qualitative results of novel view synthesis on Deblur-NeRF dataset.} We evaluate novel view synthesis performance against baseline methods under 3, 6, and 9 input view settings.
        Our approach consistently produces higher-fidelity renderings, recovering significantly more fine-grained details than competing methods.
    }
    \label{fig:deblur}
\end{figure*}

\vspace{+6pt}
\noindent \textbf{Baselines.}
We benchmark our method against a comprehensive set of state-of-the-art approaches spanning two relevant domains.
Specifically, to validate the reconstruction capability under sparse and blurry inputs, we compare our approach with BAD-Gaussians~\cite{zhao2024badgs} and Sparse-DeRF\footnote{This method is not open-source, and we reproduct it based on DP-NeRF~\cite{lee2023dp} and MPR-Net~\cite{mpr}, denoted as Sparse-DeRF$^* $.}~\cite{sparse-derf-tpami2025}.
Furthermore, to assess the performance of generative priors, we compare our method with recent 3D-aware generative models, including Difix3D+\cite{wu2025difix3d+} and GenFusion\cite{wu2025genfusion}.

\subsection{Experiment Results}
\label{sec:exps:exp_results_and_comparison}

\vspace{+6pt}
\noindent \textbf{Evaluation of Novel View Synthesis.}
We benchmark our method against state-of-the-art approaches, including GenFusion~\cite{wu2025genfusion}, Difix3D+\cite{wu2025difix3d+}, and Sparse-DeRF\cite{sparse-derf-tpami2025}.
As shown in Table~\ref{tab:exp:nvs_deblurnerf_syn_data} and Table~\ref{tab:exp:nvs_deblurnerf_real_data}, CoherentGS achieves a great performance lead on the Deblur-NeRF dataset across all sparsity settings.
Notably, in the most challenging 3-view scenario, our method outperforms GenFusion achieving a 4.28 dB gain in PSNR.
This significant improvement underscores the efficacy of our approach in handling the ill-posed problem of sparse deblurring, where traditional constraints typically collapse.

\input{Tables/table_exp_dl3dv}
\begin{figure*}[!t]
    \centering
        \includegraphics[width=0.95\textwidth]{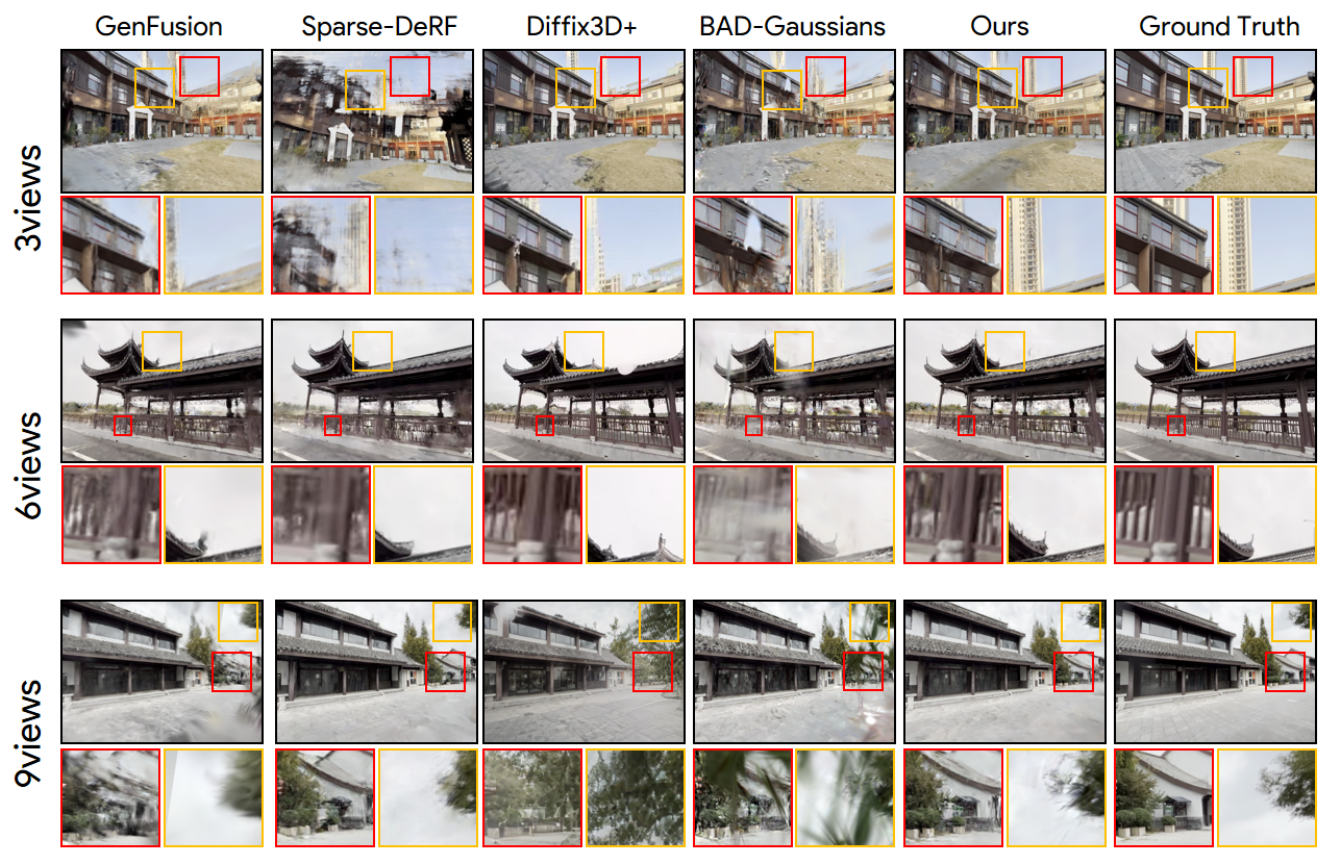}
    \centering
    \caption{
        \textbf{Qualitative results of novel view synthesis on proposed \textsc{DL3DV-Blur} Dataset.} Compared the novel view with other methods with baselines rendering quality using 3, 6, and 9 input views, our approach produces more realistic rendering results with fine-grained details.
    }
    \label{fig:dl3dv}
\end{figure*}

Qualitative results provided in Fig.~\ref{fig:deblur} highlight the superior fidelity and structural coherence of CoherentGS.
While our method reconstructs photorealistic textures with sharp edges, competing approaches suffer from distinct failure modes inherent to their paradigms.
Specifically, Sparse-DeRF struggles to resolve fine-grained geometry due to the ambiguity caused by sparse, motion-blurred inputs.
More critically, while diffusion-guided methods like Difix3D+ and GenFusion employ strong generative priors to enhance visual quality, they rely heavily on fixed camera poses estimated from degraded observations.
This dependency creates a critical bottleneck: when initial trajectories are inaccurate, the resulting reprojection misalignments are erroneously baked into the geometry by the generative prior, causing structural distortions rather than correcting them.
Furthermore, the severe blur corrupts the semantic evidence required by these priors, making diffusion models prone to hallucinating content that contradicts the underlying scene.
By jointly refining camera poses and scene representation, CoherentGS breaks this error propagation loop, yielding results that are not only visually pleasing but also geometrically faithful to the true scene.

\noindent \textbf{Evaluation of Generalization in large-scale scenes.}
To evaluate the robustness of CoherentGS in unstructured environments, we use the proposed DL3DV-Blur which is simulated motion blur on five outdoor scenes and conduct assessments under sparse-input protocols (3, 6, and 9 views).
As reported in Table~\ref{tab:exp:nvs_deblurnerf_syn_data}, CoherentGS consistently outperforms competing methods across all sparsity levels, demonstrating strong generalization capabilities in geometrically complex, unseen scenarios.
Beyond numerical improvements, our method delivers superior visual fidelity.
In contrast to generative 3DGS-based baselines like Difix3D+ and GenFusion, which tend to over-smooth fine details, CoherentGS effectively preserves high-frequency textures and intricate geometric structures.
This results in sharper object boundaries and a more faithful reconstruction of thin or cluttered elements.
Qualitative results in Fig.~\ref{fig:dl3dv} further highlight these advantages, confirming that CoherentGS produces geometrically clear and view-consistent renderings, even when optimized from sparsely sampled and severely blurred observations.

\input{Tables/table_exp_deblur_real}

\begin{figure*}[t!]
    \centering
    \includegraphics[width=1.0\linewidth]{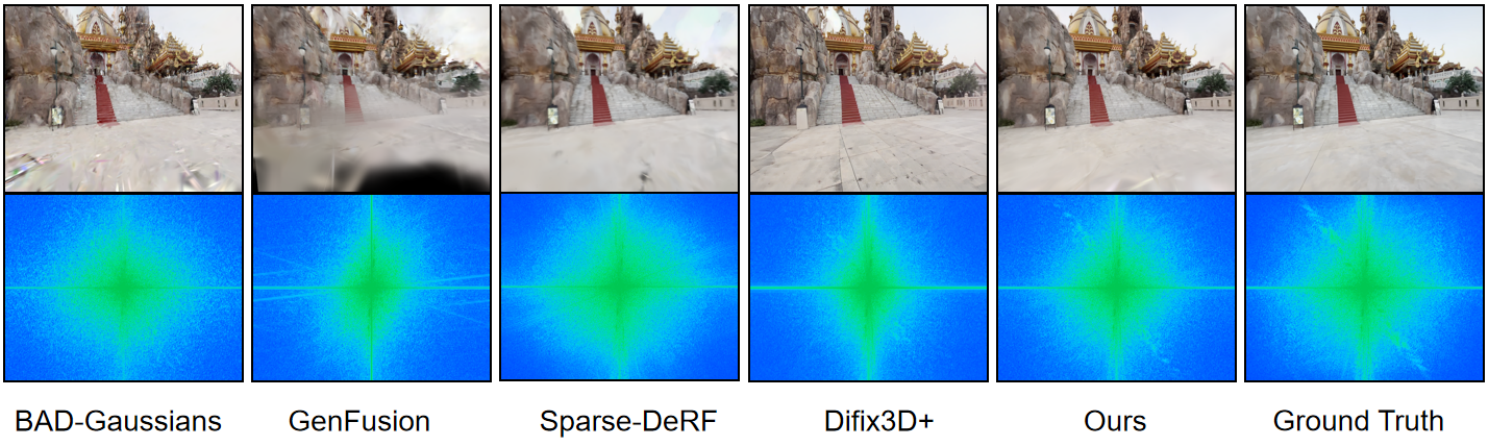}
    \caption{\textbf{Frequency Spectrum Analysis.}
    Compared to BAD-Gaussians and GenFusion, our approach produces a frequency distribution that closely aligns with the \textit{Ground Truth}. This confirms that our method effectively recovers realistic textures while suppressing the directional artifacts common in generative priors.
    }
    \label{fig:FFT}
\end{figure*}

\input{Tables/table_prior}

\vspace{+6pt}
\noindent \textbf{Spectral Analysis via Fast Fourier Transform.}
To validate structural fidelity beyond the spatial domain, we compare the spectra of BAD-Gaussians, GenFusion, Sparse-Derf,Difix3D+ and our synthesized views against the Ground Truth (GT) using 2D Fast Fourier Transform (FFT).
As shown in Fig.~\ref{fig:FFT}, the GT spectrum exhibits a compact low-frequency center that decays naturally into rich, slightly asymmetrical high-frequency structures, reflecting the scene's diverse and natural geometric details.
Our method produces a spectral profile highly congruent with the GT, preserving the scale of central energy concentration and the irregular high-frequency distribution with minimal amplitude loss.
This indicates that CoherentGS effectively suppresses blur while recovering realistic textures without introducing structural artifacts.
In contrast, the spectrum of BAD-Gaussians displays a rapid radial energy decay, indicating a severe attenuation of high-frequency components. This confirms that the optimization process over-smooths the reconstruction, failing to recover fine edges and details.
Conversely, GenFusion and Difix3D+ manifests prominent horizontal and vertical spectral spikes and distinct sidelobes.
This anisotropic energy concentration corresponds to directional grid-like artifacts rather than naturally distributed textures.
Although GenFusion exhibits strong high-frequency amplitude, its deviation from the GT pattern implies that the generated details are largely spectral hallucinations or structural noise.
Overall, our method achieves the closest frequency-domain alignment with GT, corroborating that our diffusion-guided prior maintains semantic and structural fidelity while avoiding both over-smoothing and artifact induction.

\vspace{+6pt}
\noindent \textbf{Efficiency Analysis of Training and Inference.}
To verify the computational efficiency of the proposed method, we compare the Storage consumption as well as the training and inference time of different approaches on the DL3DV-Blur dataset. As shown in Table~\ref{tab:ablation_progressive_training_scaling}, under a unified setting, the number of training iterations for Gaussian-based methods is fixed to 7000 steps, while Sparse-DeRF is trained for 200000 steps. After training, we perform a single rendering of a video sequence with 30 fps and 240 frames in total.
Compared with Bad-Gaussians, which relies solely on photometric reprojection, our CoherentGS even after introducing a diffusion prior and a deblur prior as additional supervision, only increases storage consumption to 16.8G, and the training time rises by merely 3.1 min. The inference time is 0.6 min, accounting for less than 5\% of the overall runtime. This indicates that the extra computational cost remains modest while achieving significantly improved reconstruction quality.
Compared with Difix3D+ and GenFusion, which depend on complex inference pipelines, CoherentGS further demonstrates an advantage in overall efficiency. By explicitly integrating priors during training, CoherentGS substantially simplifies the inference procedure. In contrast to Sparse-DeRF, which requires 25.6G memory and up to 132h of training, CoherentGS is one order of magnitude more efficient in terms of both memory and time, while still delivering high-quality reconstructions and being more friendly to practical computational budgets.

\input{Tables/table_training_cost}

\begin{figure}[t!]
    \centering
    \includegraphics[width=\linewidth]{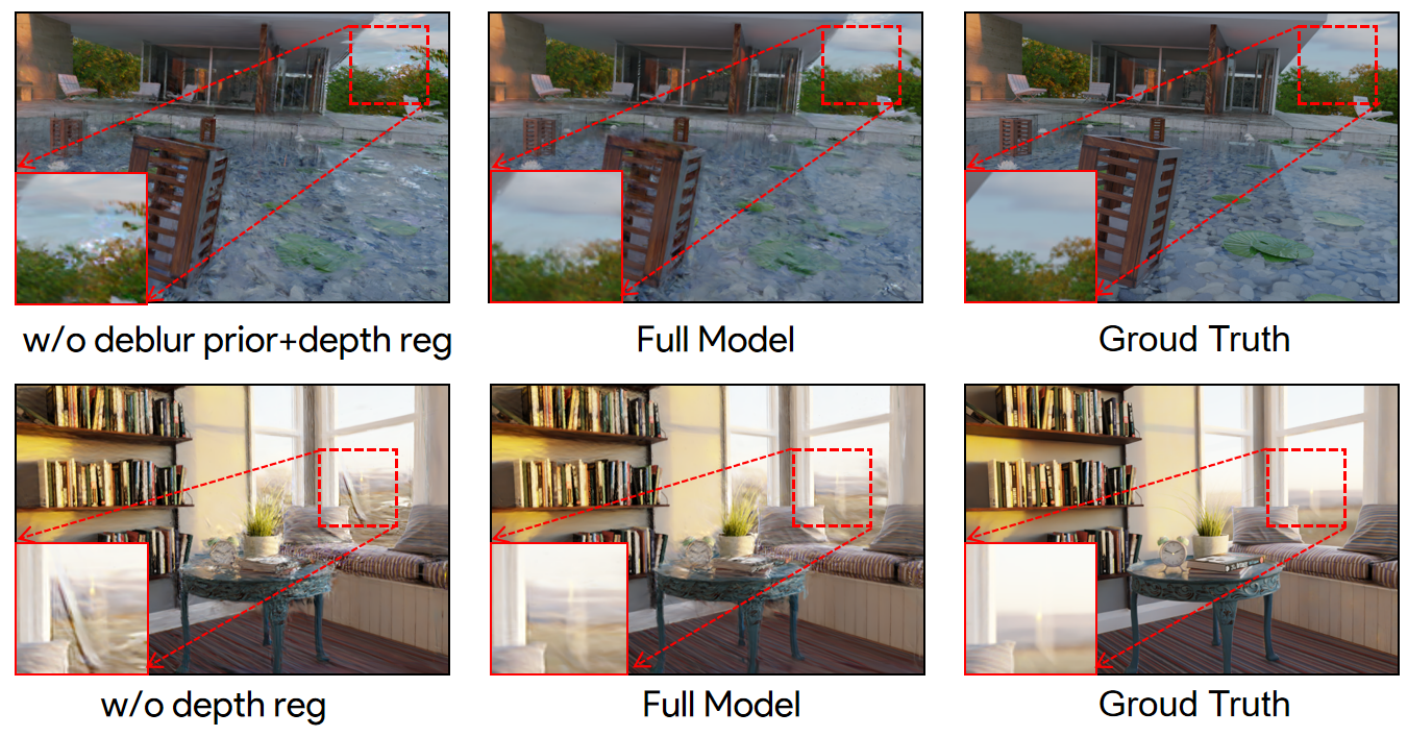}
    \caption{
     \textbf{Visual ablation of the Supervised Signals.} We compare our model against variants lacking specific components to validate their contributions.
    }
    \label{fig:prior}
\end{figure}

\subsection{Ablation Study}
\label{sec:exps:ablation}
\noindent \textbf{Effectiveness of Supervised Signals.}
To dissect the contribution of each component, we conduct an ablation study on the synthetic dataset under the 3-view setting, sequentially incorporating the diffusion prior, deblur prior, and depth regularization.We take Bad-gaussians as our baseline.
Quantitative results in Table~\ref{tab:ablation:components} show that our full model consistently yields the best performance across all metrics.
The qualitative comparisons in Fig.~\ref{fig:prior} further reveal the underlying mechanics of these components.
Specifically, removing the deblur prior significantly degrades 3D consistency.
Under sparse and blurry conditions, the geometry is inherently ambiguous. Without the deblur prior acting as a semantic anchor, the diffusion model struggles to distinguish between high-frequency textures and blur artifacts, leading to inconsistent hallucinations across views. By explicitly decoupling texture from motion blur, the deblur prior provides a cleaner guidance signal for generative refinement.
Complementing this, depth regularization proves essential for suppressing geometric noise.
Omitting this term results in severe floating artifacts and near-camera noise in novel views. A critical insight here is that since the diffusion prior primarily optimizes 2D appearance, it may satisfy visual constraints by projecting realistic textures onto erroneous geometry (e.g., floaters). The depth regularization enforces geometric smoothness, ensuring that the perceptual improvements translate into correct underlying 3D structures rather than superficial texture mapping.

\begin{figure}[t!]
    \centering
    \includegraphics[width=0.88\linewidth]{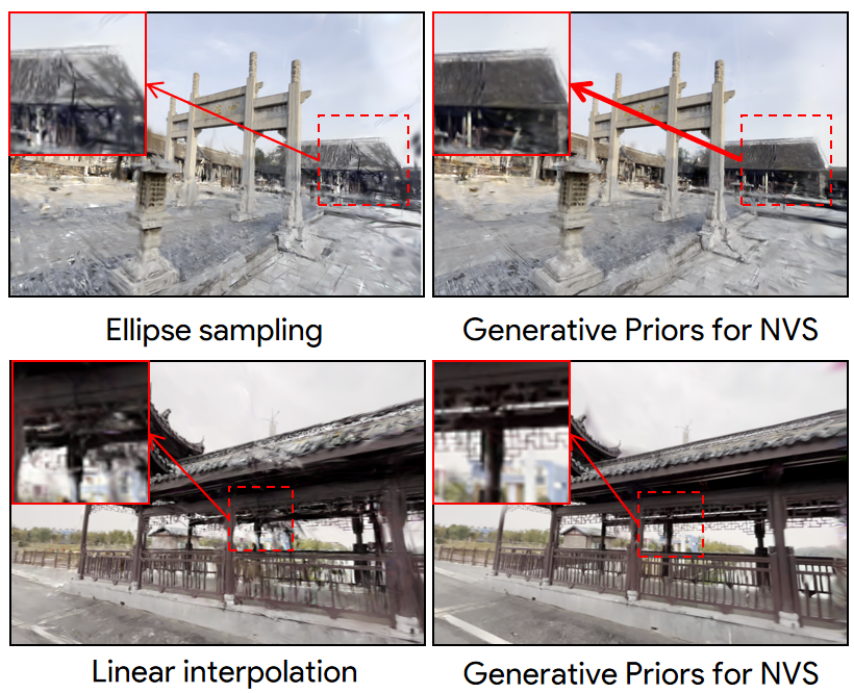}
    \caption{
     \textbf{Effectiveness of the SACN Strategy.} We compare with widely used linear interpolation and ellipse sampling. The red boxes show the most prominent differences.
    }
    \label{fig:traj}
\end{figure}

\vspace{+10pt}
\noindent \textbf{Effectiveness of SACN.}
To validate the effectiveness of our proposed Scene Adaptive Consistency Normalization, we compare it against two standard interpolation strategies: linear interpolation and elliptical trajectory interpolation.
As visualized in Fig.~\ref{fig:traj}, standard strategies suffer from inherent limitations:
Linear interpolation constrains novel views to the baseline between input frames. This strategy offers minimal information gain, as the rendered views provide limited angular variation and fail to expose occluded or under-optimized regions to the diffusion prior. Consequently, the optimization becomes inefficient due to semantic redundancies.
Conversely, elliptical interpolation maximizes angular coverage but often ventures into unobserved regions outside the visual hull of the input views. Since the diffusion prior lacks contextual image evidence in these blind spots, it tends to hallucinate content that is inconsistent with the true scene. 
Forcing the model to incorporate these erroneous priors will lead to artifact propagation and geometric distortion.

\input{Tables/table_traj}

In contrast, our strategy achieves reliability-aware exploration. 
It actively guides the camera to explore regions that are geometrically uncertain yet semantically recoverable. By maximizing the potential of the diffusion prior within a reliable trust region, our method effectively repairs artifacts while avoiding the semantic inconsistencies caused by aggressive sampling.
Quantitative results in Table \ref{tab:trajectory} confirm that this reference-guided strategy significantly improves reconstruction quality compared to topology-agnostic sampling methods.

\vspace{+10pt}
\noindent\textbf{Analysis of Warm-up Strategy.}
\begin{figure}[t!]
    \centering
    \includegraphics[width=\linewidth]{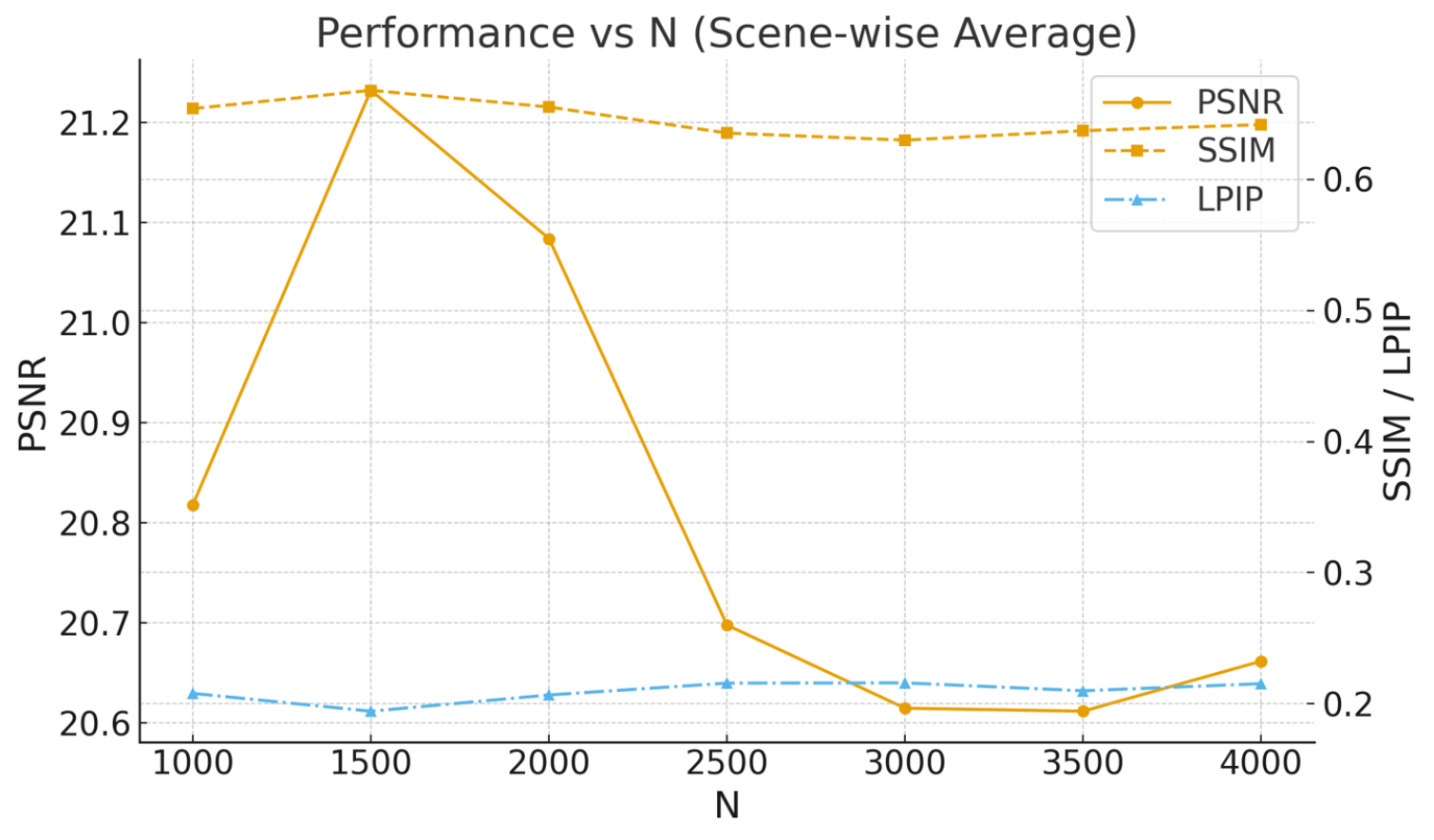}
    \caption{
     \textbf{Analysis of the Warm-up Strategy.} 
     The model achieves the best balance between geometric stability and semantic refinement at $N=1500$, yielding the highest scores.
    }
    \label{fig:warmup}
\end{figure}
To determine the optimal timing for introducing the generative prior, we evaluate performance across different warm-up iterations.
As shown in Fig.~\ref{fig:warmup}, the reconstruction quality exhibits a clear unimodal trend.
Premature Injection: Introducing the diffusion prior too early (e.g., before the deblurring module converges) is detrimental. At this stage, the scene representation has not yet disentangled motion blur from geometry. Consequently, the generative prior risks interpreting motion blur as intrinsic texture, erroneously solidifying low-frequency artifacts into the model.
Delayed Injection: Conversely, introducing the prior too late leaves an insufficient optimization window. With limited iterations remaining, the gradient guidance from the diffusion prior fails to fully propagate to unseen regions, resulting in sub-optimal refinement and limiting the quality of synthesized views.
Our empirical results suggest that waiting for the explicit deblurring module to stabilize before injecting the prior achieves the best balance.

%% file: Tables/table_exp_deblur.tex
\begin{table*}[!ht]
\centering
\caption{Quantitative comparison of novel view synthesis on the synthetic dataset~\cite{deblur-nerf}. We compare the rendering quality with baselines given 3, 6, and 9 views. Each column is colored as: \colorbox{orange!20}{best} and \colorbox{topicblue!20}{second best.}}
\label{tab:exp:nvs_deblurnerf_syn_data}
\renewcommand{\arraystretch}{1.1}
\resizebox{\linewidth}{!}{
\begin{tabular}{l|cccc cccc cccc}
\toprule
\multirow{2}{*}{Method} & \multicolumn{4}{c}{PSNR$\uparrow$} & \multicolumn{4}{c}{SSIM$\uparrow$} & \multicolumn{4}{c}{LPIPS$\downarrow$} \\
\cmidrule(lr){2-5} \cmidrule(lr){6-9} \cmidrule(lr){10-13}
& 3-view & 6-view & 9-view & Average & 3-view & 6-view & 9-view & Average & 3-view & 6-view & 9-view & Average \\
\midrule
BAD-Gaussians~\cite{zhao2024badgs} &19.58 &\colorbox{topicblue!20}{22.67} &\colorbox{topicblue!20}{25.42} &\colorbox{topicblue!20}{22.55} &0.555&\colorbox{topicblue!20}{0.727} &\colorbox{topicblue!20}{0.801} &\colorbox{topicblue!20}{0.694} &0.274 &\colorbox{topicblue!20}{0.173} &\colorbox{topicblue!20}{0.104} &\colorbox{topicblue!20}{0.183} \\
Sparse-DeRF$^*$~\cite{sparse-derf-tpami2025}   &\colorbox{topicblue!20}{19.70} &22.25 &23.25 &21.73  &\colorbox{topicblue!20}{0.561} & 0.722& 0.746&0.676 &\colorbox{topicblue!20}{0.272} & 0.169&0.143 & 0.195\\
\midrule
Difix3D+~\cite{wu2025difix3d+}       &18.43 & 19.14&19.74 & 19.11&0.504 &0.551&0.584 & 0.547& 0.307&0.287 & 0.275&0.289 \\
GenFusion~\cite{wu2025genfusion}    &16.84 &18.49 &19.47 &18.26 &0.509 &0.556 &0.600 &0.555 &0.507 &0.475 &0.452 &0.478 \\
\midrule
\textbf{Ours} & \colorbox{orange!20}{21.12} & \colorbox{orange!20}{23.87} & \colorbox{orange!20}{26.36} & \colorbox{orange!20}{23.78} & \colorbox{orange!20}{0.671} & \colorbox{orange!20}{0.783} & \colorbox{orange!20}{0.851} & \colorbox{orange!20}{0.768} & \colorbox{orange!20}{0.195} & \colorbox{orange!20}{0.122} & \colorbox{orange!20}{0.08} & \colorbox{orange!20}{0.132} \\
\bottomrule
\end{tabular}
}
\end{table*}

%% file: Tables/table_exp_dl3dv.tex
\begin{table*}[!ht]
\centering
\renewcommand{\arraystretch}{1.1}
\caption{\textbf{Quantitative comparison of novel view synthesis on proposed \textsc{DL3DV-Blur}.} We compare the rendering quality with baselines given 3, 6, and 9 views. Each column is colored as: \colorbox{orange!20}{best} and \colorbox{topicblue!20}{second best.}}
\label{tab:exp:nvs_dl3dv}
\resizebox{\linewidth}{!}{
\begin{tabular}{l|cccc cccc cccc}
\toprule
\multirow{2}{*}{Method} & \multicolumn{4}{c}{PSNR$\uparrow$} & \multicolumn{4}{c}{SSIM$\uparrow$} & \multicolumn{4}{c}{LPIPS$\downarrow$} \\
\cmidrule(lr){2-5} \cmidrule(lr){6-9} \cmidrule(lr){10-13}
& 3-view & 6-view & 9-view & Average & 3-view & 6-view & 9-view & Average & 3-view & 6-view & 9-view & Average \\
\midrule
BAD-Gaussians~\cite{zhao2024badgs}  &\colorbox{topicblue!20}{15.08} &\colorbox{topicblue!20}{18.62} &\colorbox{topicblue!20}{19.82} &\colorbox{topicblue!20}{17.84} &\colorbox{topicblue!20}{0.501} &\colorbox{topicblue!20}{0.635} &0.671 &\colorbox{topicblue!20}{0.602} &0.466 &\colorbox{topicblue!20}{0.313} &\colorbox{topicblue!20}{0.267} &\colorbox{topicblue!20}{0.349} \\
Sparse-DeRF$^*$~\cite{sparse-derf-tpami2025}   &14.84 &18.61 &19.72 &17.72 &0.483 &0.618 &\colorbox{topicblue!20}{0.674} &0.592 &0.501 & 0.325&0.277 &0.368 \\
\midrule
Difix3D+~\cite{wu2025difix3d+}       &14.36 & 15.53&16.16 &15.35 &0.492 &0.538 &0.594 &0.541 &\colorbox{topicblue!20}{0.404} & 0.416&0.336 &0.385 \\
GenFusion~\cite{wu2025genfusion}          &12.82 &16.42 &17.83 &15.69 &0.502 &0.609 &0.646 &0.586 &0.541 &0.478 &0.469 &0.496 \\
\midrule
\textbf{Ours} &\colorbox{orange!20}{17.48} &\colorbox{orange!20}{19.87} &\colorbox{orange!20}{21.64} &\colorbox{orange!20}{19.67} &\colorbox{orange!20}{0.639} &\colorbox{orange!20}{0.678} &\colorbox{orange!20}{0.731} &\colorbox{orange!20}{0.683} &\colorbox{orange!20}{0.368} &\colorbox{orange!20}{0.267} &\colorbox{orange!20}{0.233} &\colorbox{orange!20}{0.289} \\
\bottomrule
\end{tabular}
}
\end{table*}

%% file: Tables/table_exp_deblur_real.tex
\begin{table*}[!ht]
\centering
\caption{Quantitative comparison of novel view synthesis on the real-scene dataset~\cite{deblur-nerf}. We compare the rendering quality with baselines given 3, 6, and 9 views. Each column is colored as: \colorbox{orange!20}{best} and \colorbox{topicblue!20}{second best.}}
\label{tab:exp:nvs_deblurnerf_real_data}
\renewcommand{\arraystretch}{1.1}
\resizebox{\linewidth}{!}{
\begin{tabular}{l|cccc cccc cccc}
\toprule
\multirow{2}{*}{Method} & \multicolumn{4}{c}{PSNR$\uparrow$} & \multicolumn{4}{c}{SSIM$\uparrow$} & \multicolumn{4}{c}{LPIPS$\downarrow$} \\
\cmidrule(lr){2-5} \cmidrule(lr){6-9} \cmidrule(lr){10-13}
& 3-view & 6-view & 9-view & Average & 3-view & 6-view & 9-view & Average & 3-view & 6-view & 9-view & Average \\
\midrule
BAD-Gaussians~\cite{zhao2024badgs} &\colorbox{topicblue!20}{18.57} &\colorbox{topicblue!20}{20.61} &\colorbox{topicblue!20}{21.99} &\colorbox{topicblue!20}{20.39} &\colorbox{topicblue!20}{0.495} &\colorbox{topicblue!20}{0.619} &\colorbox{topicblue!20}{0.659} &\colorbox{topicblue!20}{0.591} &\colorbox{topicblue!20}{0.337} &\colorbox{topicblue!20}{0.257} &\colorbox{topicblue!20}{0.216} &\colorbox{topicblue!20}{0.270}
\\
Sparse-DeRF$^*$~\cite{sparse-derf-tpami2025}   &17.78 &20.01 &21.16 &19.65 &0.463 &0.598 &0.618 &0.560 &0.379 & 0.281&0.242 &0.300 \\
\midrule
Difix3D+~\cite{wu2025difix3d+}       &18.06 & 19.50&19.79 & 19.12&0.533 &0.574&0.585 & 0.564& 0.415&0.390 & 0.368&0.391 \\
GenFusion~\cite{wu2025genfusion}    &13.42 &15.95 &16.19 &15.18 &0.346 &0.473 &0.472&0.430 &0.588 &0.533 &0.540 &0.554 \\
\midrule
\textbf{Ours} & \colorbox{orange!20}{19.57} & \colorbox{orange!20}{22.01} & \colorbox{orange!20}{23.44} & \colorbox{orange!20}{21.67} & \colorbox{orange!20}{0.575} & \colorbox{orange!20}{0.681} & \colorbox{orange!20}{0.747} & \colorbox{orange!20}{0.668} & \colorbox{orange!20}{0.297} & \colorbox{orange!20}{0.201} & \colorbox{orange!20}{0.167} & \colorbox{orange!20}{0.221} \\
\bottomrule
\end{tabular}
}
\end{table*}

%% file: Tables/table_prior.tex
\begin{table}[t!]
\centering
\caption{Ablation study of our method's components.}
\label{tab:ablation:components}
\begin{tabular}{lccc}
\toprule
\textbf{CoherentGS} & PSNR$\uparrow$ & SSIM$\uparrow$ & LPIPS$\downarrow$ \\
\midrule
\makecell[l]{\textbf{baseline}} & 19.58& 0.555&0.274 \\
+Geometric Priors & 20.59&0.615 &0.218 \\
+Deblurring Priors & 20.83&0.647 &0.207 \\
+Depth Loss &\textbf{21.12} &\textbf{0.671} &\textbf{0.195} \\

\bottomrule
\end{tabular}
\vspace{-10pt}
\end{table}

%% file: Tables/table_training_cost.tex
\begin{table}[h]
\centering
\begin{tabular}{l|c|c|c}
\toprule
\textbf{Method} & \textbf{Storage} & \textbf{Training Time} & \textbf{Inference Time} \\
\midrule
(a) Bad-Gaussians &9.63G & 8.6min & 0.5min\\
(b) Difix3D+ &14.9G & 3.7min & 6.9min \\
(c) Genfusion &20.2G & 3.9min & 9.19min \\
(d) Sparse-DeRF &25.6G & 132h &1.8min \\
(e) Ours &16.8G & 11.7min &0.6min  \\
\bottomrule
\end{tabular}
\caption{
Comparison of GPU storage usage, training time, and inference time of different methods on the DL3DV-Blur dataset. CoherentGS maintains competitive storage and training cost while significantly reducing inference overhead.
}
\label{tab:ablation_progressive_training_scaling}
\end{table}

%% file: Tables/table_traj.tex
\begin{table}[t!]
  \centering
  \caption{Ablation study on different camera trajectories.}
  \label{tab:trajectory}
  \resizebox{0.8\linewidth}{!}{
  \begin{tabular}{lccc}
    \toprule
    \textbf{Trajectory} & PSNR$\uparrow$ & SSIM$\uparrow$ & LPIPS$\downarrow$ \\
    \midrule
    Linear interpolation   & 17.24 & 0.575 & 0.372 \\
    Ellipse sampling       & 15.64 & 0.550 & 0.428 \\
    SACN & \textbf{17.48} & \textbf{0.639} & \textbf{0.368} \\
    \bottomrule
  \end{tabular}
  }
\end{table}

%% file: Sections/5_limitation_and_conclusion.tex
\section{Conclusion and Limitations}
\noindent \textbf{Conclusion.}
In this paper, we present CoherentGS, a novel 3D Gaussian Splatting framework designed to achieve high-fidelity 3D reconstruction from sparse and motion-blurred inputs. 
We identify that these prevalent real-world degradations create a vicious cycle where sparse views impede blur resolution, and blur degrades high-frequency details essential for view alignment, leading to fragmented reconstructions and low-frequency bias.
To effectively break this cycle, our core contribution is a synergistic dual-prior strategy. 
This strategy intelligently integrates a specialized deblurring network for robust photometric guidance and a powerful diffusion model providing geometric priors for scene completion. These are further supported by a consistency-guided camera exploration module for adaptive viewpoint planning and a depth regularization loss for geometric plausibility.
Extensive quantitative and qualitative experiments on both synthetic and real-world scenes, utilizing as few as 3, 6, and 9 input views, unequivocally demonstrate CoherentGS's superior performance. 
Our method consistently outperforms existing state-of-the-art approaches, delivering significantly more coherent, detailed, and visually realistic novel view syntheses under challenging conditions. CoherentGS thus establishes a new benchmark for robust 3D reconstruction from degraded inputs, substantially expanding the practical applicability of 3DGS.

\vspace{+10pt}
\noindent \textbf{Limitations.}
While our proposed CoherentGS significantly advances 3D reconstruction from sparse and motion-blurred inputs, its current design focuses primarily on these specific degradations. 
Our framework is not yet optimized to robustly handle other common real-world challenges, such as defocus blur, overexposed or dark images, and complex degraded effects. 
Addressing these additional types of degraded inputs, potentially through the integration of more generalized priors or multi-degradation modeling, presents a promising avenue for future work to further broaden CoherentGS's applicability in diverse and challenging scenarios.

%% file: Sections/X_supplementary.tex
\subsection{Dataset Details}
\label{app:dataset_details}

\noindent \textbf{\textsc{DL3DV-Blur} Dataset.}
To rigorously assess the generalization capability of CoherentGS in complex, unconstrained outdoor environments, we establish a new benchmark named \textsc{DL3DV-Blur}, derived from five diverse scenes within the DL3DV-10K dataset~\cite{ling2024dl3dv}.
In constructing this dataset, we strictly adhere to the \textit{high-fidelity blur generation protocol} proposed in DAVANet~\cite{zhou2019davanet}.
As analyzed in~\cite{zhou2019davanet}, synthesizing motion blur by simply averaging frames from low-fps video sequences is insufficient to approximate realistic long-exposure photography. Such low-temporal-resolution accumulation inevitably introduces discontinuous ghosting artifacts and temporal aliasing, which differ significantly from the continuous integral of light on a camera sensor.

To address this and simulate physically realistic blur, we adopt the following pipeline:
\begin{enumerate}
    \item \textbf{Preprocessing for Diffusion Alignment:} First, to ensure compatibility with the spatial alignment requirements of the latent diffusion model, we resize the original video resolution from $960 \times 540$ to $960 \times 536$.
    \item \textbf{High-Frame-Rate Interpolation:} Following the strategy in~\cite{zhou2019davanet}, we employ the high-quality frame interpolation method SepConv-Slomo~\cite{Niklaus_ICCV_2017} to temporally upsample the source footage to 480 fps. This dense temporal sampling is crucial for filling the gaps between frames, thereby suppressing discrete artifacts.
    \item \textbf{Blur Generation via Accumulation:} We generate the final motion-blurred images by mathematically approximating the exposure integration process. This is achieved by averaging a sliding window of varying sizes (specifically 6 and 10 frames from the interpolated sequence) centered on the corresponding sharp ground truth frame.
\end{enumerate}
This rigorous process ensures that \textsc{DL3DV-Blur} exhibits smooth, and realistic motion trails, providing a challenging and reliable benchmark for evaluating sparse-view deblurring performance. The dataset is available at \href{https://huggingface.co/datasets/Passwerob/CoherentGS/tree/main}{huggingface}.

\vspace{+10pt}
\noindent \textbf{\textsc{Deblur-NeRF} Dataset.}
To comprehensively evaluate our method, we conduct experiments on the standard \textsc{Deblur-NeRF} benchmark\cite{deblur-nerf}, which comprises both synthetic and real-world subsets tailored for varying degrees of motion blur.

\begin{itemize}
\item \textbf{Synthetic Scenes:} This subset includes five diverse 3D scenes: \textit{Cozyroom}, \textit{Factory}, \textit{Pool}, \textit{Tanabata}, and \textit{Trophy}. These scenes are rendered using Blender, where photorealistic motion blur is physically simulated by accumulating multiple sub-frames along a continuous camera trajectory during the exposure interval. This process ensures that the synthesized blur strictly adheres to the physical image formation model, providing high-quality blurry inputs paired with corresponding sharp Ground Truth for quantitative evaluation.

\item \textbf{Real-world Scenes:} To assess generalization in unconstrained environments, the dataset provides 10 real-world sequences captured with a handheld camera. These sequences feature complex, non-uniform motion blur caused by natural camera shake and varying lighting conditions. Since pixel-aligned ground truth is unavailable for these in-the-wild captures, we utilize them primarily for qualitative analysis to verify the robustness of our method against real-world degradation.

\end{itemize}

\subsection{Training Details}
Our framework is implemented in PyTorch, building upon the official codebase of BAD-Gaussians~\cite{zhao2024badgs} and integrating the pre-trained prior from Difix3D+~\cite{wu2025difix3d+}.

\vspace{+6pt}
\noindent \textbf{Optimization.} 
We employ the Adam optimizer for all learnable parameters.
The learning rate schedules and densification strategies for 3D Gaussian primitives strictly follow the default configuration in~\cite{kerbl3Dgaussians,zhao2024badgs}.
For the camera trajectory modeling (Eq.~\eqref{eq:exposure_pose}), the learning rates for translation and rotation ($T_{\text{start}}, T_{\text{end}}$) are initialized at $5 \times 10^{-3}$ and exponentially decayed to $5 \times 10^{-5}$.
Regarding the loss terms, we set the blurry weight and depth regularization weight to $\lambda_{\text{1}} = 0.8, \lambda_{\text{D-SSIM}} = 0.2, \lambda_{\text{reg}} = 0.1$.The weights for the deblurring ($\lambda_{\text{pr}}$) and geometric ($\lambda_{\text{geo}}$) priors are set to 0.01.

\vspace{+6pt}
\noindent \textbf{Reliability Thresholds.} To adapt to different scene distributions, the confidence thresholds for the score model are calibrated per dataset: we set $\{s_{\max}, s_{\min}\}$ to $\{14.5, 4.5\}$ for the outdoor DL3DV-Blur dataset, and $\{8.5, 2.5\}$ for the synthetic Deblur-NeRF dataset.

\vspace{+6pt}
\noindent \textbf{Training Protocol.} The training spans a total of 7,000 iterations.
The process initiates with a warm-up phase of 1500 iterations, focusing on stabilizing the deblurring model and camera poses.
Subsequently, we activate the generative branch: the Difix prior is queried every 200 iterations to guide the optimization of under-reconstructed regions.
All experiments are conducted on a single NVIDIA A6000 with 48GB memory, taking approximately 12.3 minutes per scene.

\vspace{+6pt}
\noindent \textbf{Training Scene Indices.} 
In our training pipeline, we adopt fixed viewpoint selection settings for both the Deblur-NeRF dataset and our proposed dataset. 
The specific image indices used to train \textit{CoherentGS} are detailed in Table~\ref{tab:app:traing indices}. 
Notably, the indices for the 3-view and 6-view settings are constructed as subsets of the 9-view configuration.
\input{Tables/table_training_indices}

\subsection{Network Architectures of Pre-trained Priors}
\label{sec:app:pretained_priors}
To facilitate robust reconstruction under sparse and degraded observations, our framework integrates two specialized pre-trained models. Here, we detail their architectural designs and specific configurations used in our pipeline.

\subsubsection{Deblurring Prior: EVSSM}
For the task of recovering sharp semantic cues from motion-blurred inputs, we adopt the Efficient Visual State Space Model (EVSSM)~\cite{EVSSM} as our deblurring backbone. Unlike traditional CNN-based methods, EVSSM leverages the linear complexity of State Space Models (SSMs) to efficiently model long-range dependencies, which is critical for restoring large-scale motion blur patterns.

\begin{figure*}[t!]
    \centering
    \includegraphics[width=0.9\linewidth]{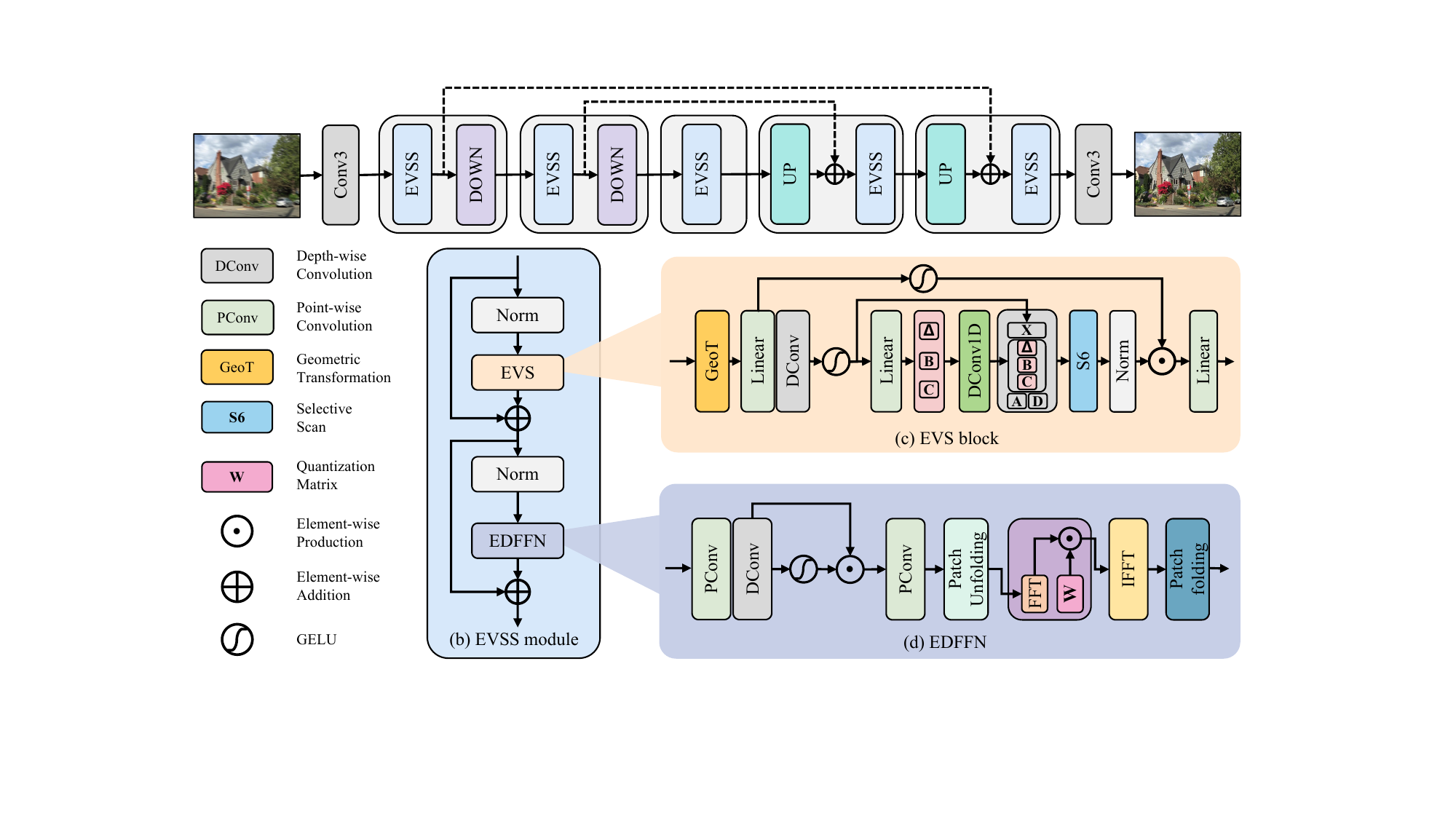}
    \caption{\textbf{The pipeline of ESSVM}~\cite{EVSSM}.
    }
    \label{fig:app:evssm_pipeline}
\end{figure*}

As shown in Fig.\ref{fig:app:evssm_pipeline}, the architecture comprises two core components designed to maximize information restoration while minimizing computational overhead:
\begin{itemize}
    \item \textbf{Visual Scanning \& Geometric Transformation:} To capture non-local features across different orientations, the input features $F_{\text{in}}$ undergo geometric transformations before being processed by SSM core. This multi-directional scanning strategy enhances receptive field without increasing parameter count.
    \item \textbf{Frequency-Aware Filtering:} Recognizing that blur predominantly affects high-frequency components, EVSSM incorporates a Frequency EDFFN module. Features are transformed via Fourier Transform $\mathcal{F}(\cdot)$, allowing the network to explicitly filter and restore spectral components:
    \begin{equation}
        F_{\text{freq}} = \mathcal{F}(F_{\text{in}}), \quad F_{\text{out}} = \mathcal{Q}(F_{\text{freq}}),
    \end{equation}
    where $\mathcal{Q}(\cdot)$ denotes the learnable frequency domain filter.
\end{itemize}
The core temporal modeling is governed by the recursive SSM equations, allowing effective propagation of sharp details:
\begin{equation}
    h(t) = \mathbf{A} h(t-1) + \mathbf{B} x(t), \quad y(t) = \mathbf{C} h(t) + \mathbf{D} x(t),
\end{equation}
where $\mathbf{A}, \mathbf{B}, \mathbf{C}, \mathbf{D}$ are learnable parameters. The final deblurred output is obtained via a residual connection: $I_{\text{deblur}} = I_{\text{blur}} + \mathcal{R}(I_{\text{blur}})$. We utilize the pre-trained weights provided by~\cite{EVSSM}, which have demonstrated superior performance on dynamic scene deblurring.

\subsubsection{Generative Prior: Difix}
\begin{figure*}[t!]
    \centering
    \includegraphics[width=0.9\linewidth]{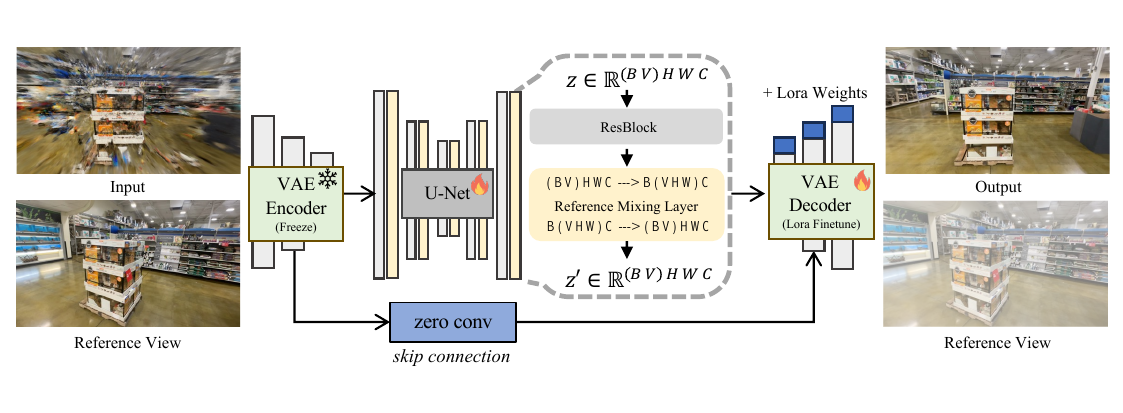}
    \caption{\textbf{The pipeline of Difix3D+}~\cite{wu2025difix3d+}.
    }
    \label{fig:app:difix3d+_pipeline}
\end{figure*}

To rectify geometric artifacts and hallucinate missing textures in sparse-view settings, we employ the DIFIX model from Difix3D+~\cite{wu2025difix3d+} as our generative prior. 
Built upon the computationally efficient SD-Turbo~\cite{sauer2025adversarial}, DIFIX is specifically fine-tuned to balance artifact removal with structural fidelity.
As shown in Fig.\ref{fig:app:difix3d+_pipeline} ,the key adaptations in DIFIX for 3D-consistent enhancement include:
\begin{itemize}
    \item \textbf{Reference-Conditioned Attention:} To ensure that the generated content remains consistent with the existing 3D scene, DIFIX injects information from clean reference views. This is achieved by concatenating the latent representations of target view $\tilde{I}$ and reference views $I_{\text{ref}}$:
    \begin{equation}
        z_{\text{joint}} = \mathcal{E}(\tilde{I} \oplus I_{\text{ref}}) \in \mathbb{R}^{V \times C \times H \times W},
    \end{equation}
    where $\mathcal{E}$ is the encoder and $\oplus$ denotes concatenation. A self-attention mechanism is then applied to $z_{\text{joint}}$ to capture cross-view dependencies, enabling the model to borrow sharp details from reference viewpoints.
    \item \textbf{Conservative Noise Level:} Instead of full stochastic generation, DIFIX operates at a reduced noise level ($\tau=200$). This design choice is crucial for our framework: it allows the model to function as a soft refiner, removing floaters and high-frequency artifacts—without deviating excessively from the original scene layout or introducing uncontrolled hallucinations.
\end{itemize}

\subsection{Additional Experiment Results}
\label{sec:app:add_exps}
\noindent \textbf{Per-scene Quantitative Breakdown}
In this section, we provide a comprehensive breakdown of the quantitative evaluation presented in the main manuscript.
While the main text reports aggregated metrics averaged across the \textsc{Deblur-NeRF} and \textsc{DL3DV-Blur} datasets in Table~\ref{tab:exp:nvs_deblurnerf_syn_data}, Table~\ref{tab:exp:nvs_dl3dv}, and Table~\ref{tab:exp:nvs_deblurnerf_real_data}, this supplementary section details the scene-wise performance.
Table~\ref{app:tab:3views} through Table~\ref{app:tab:dl3dv_blur_9views} present individual results for each scene under sparse input settings .
As evidenced by these detailed comparisons, \textbf{CoherentGS} achieves the best quantitative performance in the majority of scenes.
It is worth noting that while generative baselines like Difix3D+ occasionally yield competitive scores in specific perceptual metrics, they often suffer from severe view-inconsistency artifacts. Standard 2D metrics of novel view synthesis applied to individual frames may not fully penalize such geometric discrepancies.
In contrast, our method maintains superior 3D consistency and structural fidelity while effectively removing motion blur, demonstrating robust performance across diverse scenarios.

\begin{figure*}[t!]
    \centering
    \includegraphics[width=1.0\linewidth]{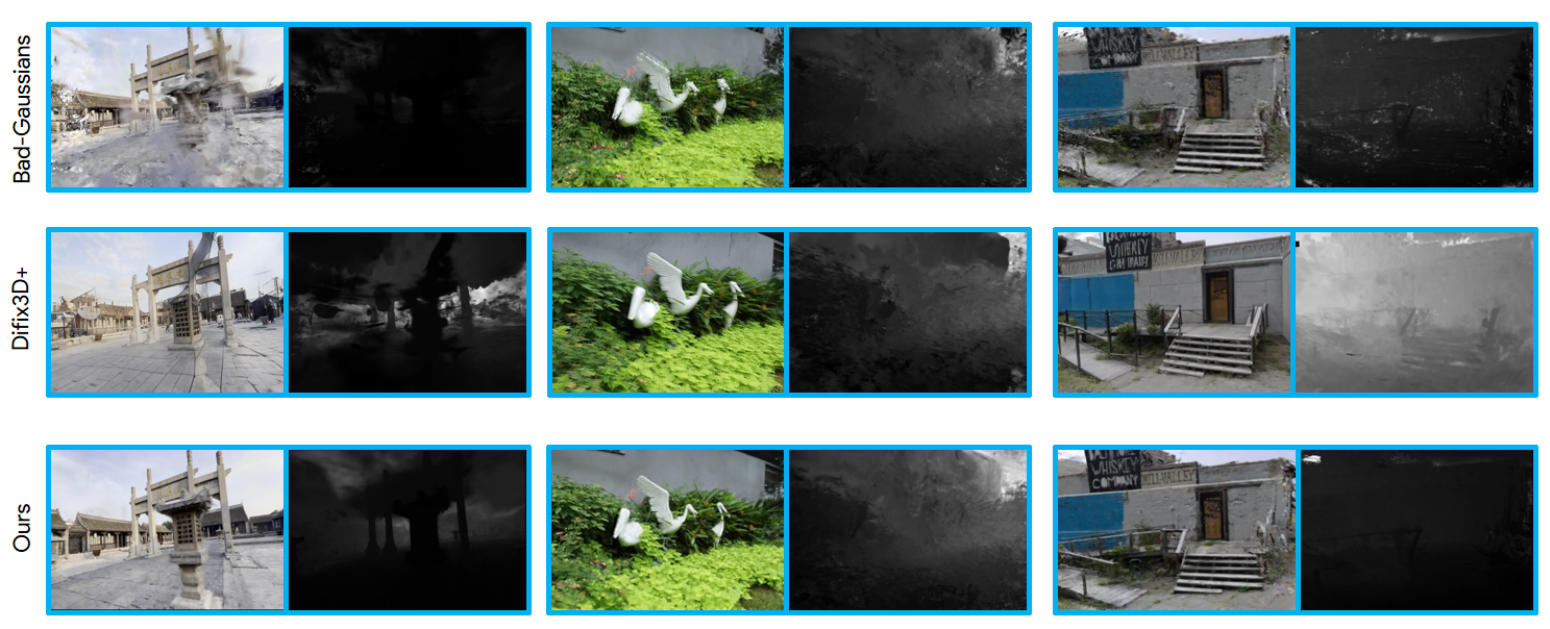}
    \caption{\textbf{Depth and Geometry Analysis.}
    Compared to BAD-Gaussians and Difix3D+, our approach produces reconstructions with significantly improved multi-view geometric consistency, recovering coherent 3D structure while suppressing the directional artifacts commonly introduced by generative priors.
    }
    \label{fig:depth}
\end{figure*}

\vspace{+6pt}
\noindent \textbf{Geometry Analysis via Rendered Depthmap.}
To validate the structural consistency of our method, we compare the depth maps of novel views generated by BAD-Gaussians, Difix3D+, and our approach. As shown in Fig.\ref{fig:depth}, our method produces a more continuous and monotonic depth distribution, with a clear separation between foreground and background, and smooth depth gradients along object surfaces. This indicates that the camera trajectory and 3D geometry are jointly estimated in a more stable manner within a unified framework. In contrast, the depth maps of BAD-Gaussians exhibit globally low contrast, with large regions collapsing to nearly a single depth plane and distant structures being overly smoothed, reflecting that under severe motion blur, relying solely on photometric reprojection leads the geometric solution to degenerate into a low-frequency, blurry structure. Although Difix3D+ is able to preserve part of the foreground contours, its depth maps show pronounced speckle-like noise and discontinuous local fluctuations. This arises from performing de-artifacting independently in the 2D image domain without explicit multi-view geometric consistency constraints, making it difficult for the diffusion model’s per-view textures to remain structurally and geometrically consistent when lifted into 3D space. Overall, this depth comparison further confirms that our diffusion-guided prior effectively maintains global geometric consistency and stability while avoiding both over-smoothing and artifact injection.

\begin{figure*}[t!]
    \centering
    \includegraphics[width=1.0\linewidth]{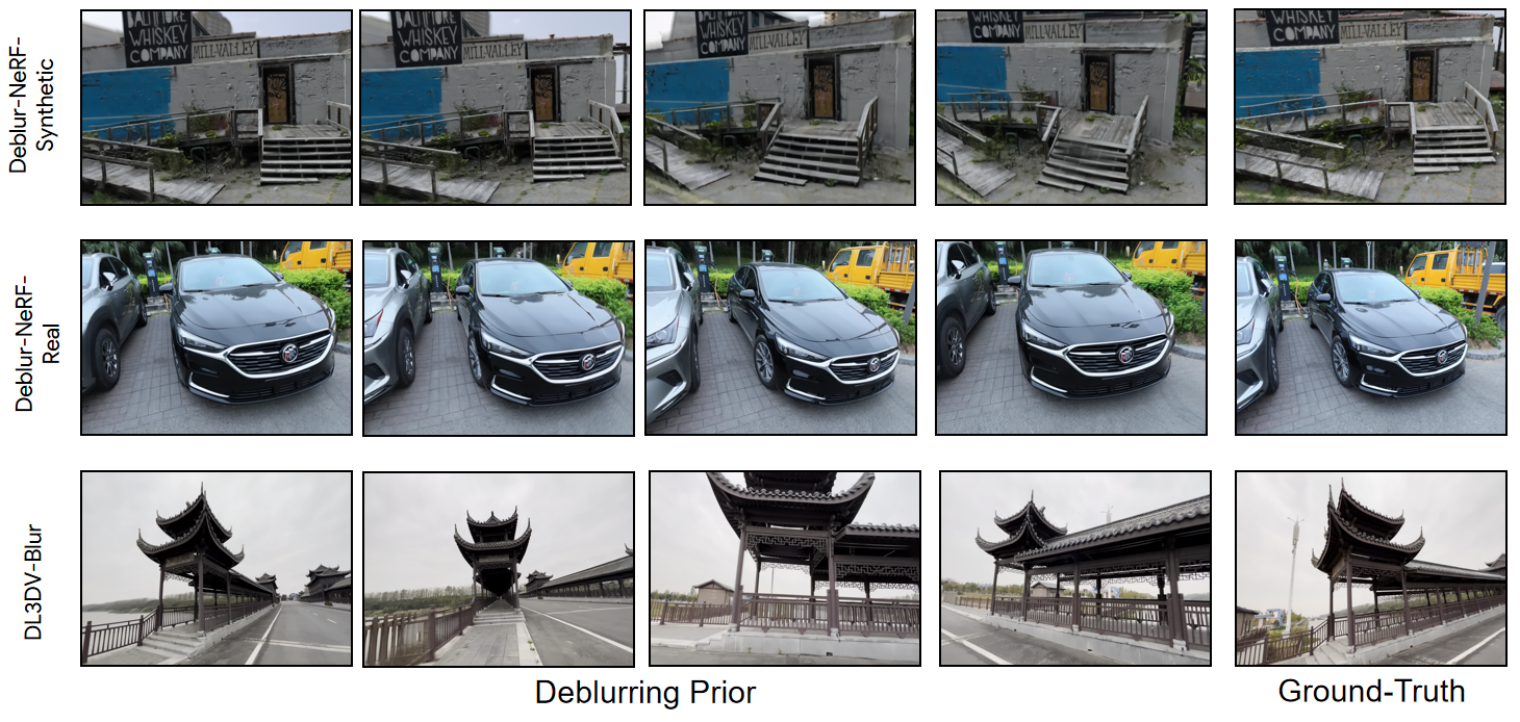}
    \caption{Visualization of the deblurring prior on Deblur-NeRF-Synthetic, Deblur-NeRF-Real, and DL3DV-Blur. The deblurring prior sharpens textures and edges while maintaining consistent multi-view geometry.
    }
    \label{fig:deblurprior}
\end{figure*}

\vspace{+6pt}
\noindent \textbf{Geometric Consistency of the Deblurring Prior}
Fig.\ref{fig:deblurprior} shows the visualization of the deblurring prior on the Deblur-NeRF-Synthetic, Deblur-NeRF-Real, and DL3DV-Blur datasets. 
For each dataset, the first four images in each row are deblurred views produced by EVSSM, and the last image is the corresponding sharp ground-truth view. 
We observe that the deblurring prior mainly enhances texture and edge details, while the object contours, occlusion relationships, and overall scene layout across different views remain consistent with the ground truth, without artifacts. 
Overall,the results on these three datasets indicate that the deblurring prior significantly improves image sharpness without breaking multi-view geometric consistency.

\begin{table*}[h]
    \centering
    \footnotesize
    \setlength{\tabcolsep}{1.4pt}
    \renewcommand\arraystretch{1.2}
\resizebox{\linewidth}{!}{
    
    \begin{tabular}{@{}lccccccccccccccc|ccc@{}}
    \toprule
        \multirow{2}{*}{Methods} & \multicolumn{3}{c}{blurcozy2room} & \multicolumn{3}{c}{blurfactory} & \multicolumn{3}{c}{blurpool} & \multicolumn{3}{c}{blurtanabata} & \multicolumn{3}{c}{blurwine} & \multicolumn{3}{c}{Average} \\ 
        \cmidrule(r){2-4} \cmidrule(r){5-7} \cmidrule(r){8-10} \cmidrule(r){11-13} \cmidrule(r){14-16} \cmidrule(r){17-19}
        & \scriptsize PSNR$\uparrow$ & \scriptsize SSIM$\uparrow$ & \scriptsize LPIPS$\downarrow$
        & \scriptsize PSNR$\uparrow$ & \scriptsize SSIM$\uparrow$ & \scriptsize LPIPS$\downarrow$
        & \scriptsize PSNR$\uparrow$ & \scriptsize SSIM$\uparrow$ & \scriptsize LPIPS$\downarrow$
        & \scriptsize PSNR$\uparrow$ & \scriptsize SSIM$\uparrow$ & \scriptsize LPIPS$\downarrow$
        & \scriptsize PSNR$\uparrow$ & \scriptsize SSIM$\uparrow$ & \scriptsize LPIPS$\downarrow$
        & \scriptsize PSNR$\uparrow$ & \scriptsize SSIM$\uparrow$ & \scriptsize LPIPS$\downarrow$ \\ 
        \midrule
        BAD-Gaussians~\cite{zhao2024badgs}
        & \colorbox{topicblue!20}{22.21} & \colorbox{topicblue!20}{0.673} & \colorbox{topicblue!20}{0.233}
        & 19.90 & 0.652 & 0.208
        & 22.67 & 0.545 & 0.241
        & 16.42 & 0.428 & 0.373
        & \colorbox{topicblue!20}{16.70} & 0.477 & 0.318
        & 19.58 & 0.555 & 0.274 \\

        Sparse-DeRF$^*$~\cite{sparse-derf-tpami2025}
        & 21.67 & 0.642 & 0.253
        & \colorbox{topicblue!20}{20.48} & \colorbox{topicblue!20}{0.701} & \colorbox{topicblue!20}{0.185}
        & \colorbox{topicblue!20}{23.11} & \colorbox{topicblue!20}{0.566} & \colorbox{topicblue!20}{0.218}
        & \colorbox{topicblue!20}{17.12} & 0.447 & 0.381
        & 16.15 & 0.451 & 0.323
        & \colorbox{topicblue!20}{19.70} & \colorbox{topicblue!20}{0.561} & \colorbox{topicblue!20}{0.272} \\

        Difix3D+~\cite{wu2025difix3d+}
        & 20.57 & 0.653 & 0.246
        & 18.31 & 0.479 & 0.335
        & 21.11 & 0.505 & 0.299
        & 15.89 & 0.426 & \colorbox{topicblue!20}{0.339}
        & 16.24 & 0.458 & \colorbox{topicblue!20}{0.317}
        & 18.46 & 0.504 & 0.307 \\

        GenFusion~\cite{wu2025genfusion}
        & 19.75 & 0.626 & 0.436
        & 16.65 & 0.545 & 0.472
        & 16.42 & 0.438 & 0.571
        & 15.21 & \colorbox{topicblue!20}{0.449} & 0.556
        & 16.17 & \colorbox{topicblue!20}{0.492} & 0.505
        & 16.84 & 0.510 & 0.508 \\
        \midrule
        \textbf{Ours}
        & \colorbox{orange!20}{\textbf{24.50}}  & \colorbox{orange!20}{\textbf{0.786}} & \colorbox{orange!20}{\textbf{0.131}}
        & \colorbox{orange!20}{\textbf{20.94}} & \colorbox{orange!20}{\textbf{0.724}} & \colorbox{orange!20}{\textbf{0.174}}
        & \colorbox{orange!20}{\textbf{23.24}} & \colorbox{orange!20}{\textbf{0.638}} & \colorbox{orange!20}{\textbf{0.210}}
        & \colorbox{orange!20}{\textbf{18.58}} & \colorbox{orange!20}{\textbf{0.611}} & \colorbox{orange!20}{\textbf{0.231}}
        & \colorbox{orange!20}{\textbf{18.37}} & \colorbox{orange!20}{\textbf{0.598}} & \colorbox{orange!20}{\textbf{0.230}}
        & \colorbox{orange!20}{\textbf{21.13}} & \colorbox{orange!20}{\textbf{0.671}} & \colorbox{orange!20}{\textbf{0.195}} \\
        \bottomrule
    \end{tabular}}
    \caption{\textbf{Quantitative results of novel view synthesis on the synthetic dataset with 3 views.} The per-scene results sum to the average values provided.Each column is colored as: \colorbox{orange!20}{best} and \colorbox{topicblue!20}{second best}.}
    \label{app:tab:3views}
\end{table*}

\begin{table*}[h]
    \centering
    \footnotesize
    \setlength{\tabcolsep}{1.4pt}
    \renewcommand\arraystretch{1.2}
\resizebox{\linewidth}{!}{
    \begin{tabular}{@{}lcccccccccccccccccc@{}}
    \toprule
        \multirow{2}{*}{Methods} & \multicolumn{3}{c}{blurcozy2room} & \multicolumn{3}{c}{blurfactory} & \multicolumn{3}{c}{blurpool} & \multicolumn{3}{c}{blurtanabata} & \multicolumn{3}{c}{blurwine} & \multicolumn{3}{c}{Average} \\ 
        \cmidrule(r){2-4} \cmidrule(r){5-7} \cmidrule(r){8-10} \cmidrule(r){11-13} \cmidrule(r){14-16} \cmidrule(r){17-19}
        & \scriptsize PSNR$\uparrow$ & \scriptsize SSIM$\uparrow$ & \scriptsize LPIPS$\downarrow$
        & \scriptsize PSNR$\uparrow$ & \scriptsize SSIM$\uparrow$ & \scriptsize LPIPS$\downarrow$
        & \scriptsize PSNR$\uparrow$ & \scriptsize SSIM$\uparrow$ & \scriptsize LPIPS$\downarrow$
        & \scriptsize PSNR$\uparrow$ & \scriptsize SSIM$\uparrow$ & \scriptsize LPIPS$\downarrow$
        & \scriptsize PSNR$\uparrow$ & \scriptsize SSIM$\uparrow$ & \scriptsize LPIPS$\downarrow$
        & \scriptsize PSNR$\uparrow$ & \scriptsize SSIM$\uparrow$ & \scriptsize LPIPS$\downarrow$ \\ 
        \midrule
        BAD-Gaussians~\cite{zhao2024badgs}
        & \colorbox{topicblue!20}{25.61} & \colorbox{topicblue!20}{0.795} & \colorbox{topicblue!20}{0.138}
        & 22.62 & 0.795 & 0.121
        & \colorbox{topicblue!20}{25.64} & 0.691 & 0.161
        & 19.09 & 0.587 & 0.261
        & \colorbox{topicblue!20}{20.43} & \colorbox{topicblue!20}{0.668} & \colorbox{topicblue!20}{0.184}
        & \colorbox{topicblue!20}{22.68} & 0.707 & 0.173 \\

        Sparse-DeRF$^*$~\cite{sparse-derf-tpami2025}
        & 23.77 & 0.754 & 0.168
        & \colorbox{topicblue!20}{22.98} & \colorbox{topicblue!20}{0.815} & \colorbox{topicblue!20}{0.109}
        & 25.32 & \colorbox{topicblue!20}{0.694} & \colorbox{topicblue!20}{0.153}
        & \colorbox{topicblue!20}{19.83} & \colorbox{topicblue!20}{0.708} & \colorbox{topicblue!20}{0.221}
        & 19.36 & 0.638 & 0.196
        & 22.25 & \colorbox{topicblue!20}{0.722} & \colorbox{topicblue!20}{0.169} \\

        Difix3D+~\cite{wu2025difix3d+}
        & 21.43 & 0.695 & 0.210
        & 19.05 & 0.528 & 0.314
        & 21.23 & 0.566 & 0.275
        & 16.62 & 0.466 & 0.335
        & 17.41 & 0.500 & 0.299
        & 19.15 & 0.551 & 0.287 \\

        GenFusion~\cite{wu2025genfusion}
        & 20.97 & 0.683 & 0.386
        & 19.63 & 0.605 & 0.447
        & 17.52 & 0.513 & 0.513
        & 17.27 & 0.476 & 0.535
        & 17.07 & 0.507 & 0.498
        & 18.49 & 0.557 & 0.476 \\
        \midrule
        \textbf{Ours}
        & \colorbox{orange!20}{\textbf{26.88}} & \colorbox{orange!20}{\textbf{0.843}} & \colorbox{orange!20}{\textbf{0.101}}
        & \colorbox{orange!20}{\textbf{23.03}} & \colorbox{orange!20}{\textbf{0.842}} & \colorbox{orange!20}{\textbf{0.090}}
        & \colorbox{orange!20}{\textbf{26.23}} & \colorbox{orange!20}{\textbf{0.735}} & \colorbox{orange!20}{\textbf{0.150}}
        & \colorbox{orange!20}{\textbf{21.25}} & \colorbox{orange!20}{\textbf{0.736}} & \colorbox{orange!20}{\textbf{0.151}}
        & \colorbox{orange!20}{\textbf{21.99}} & \colorbox{orange!20}{\textbf{0.762}} & \colorbox{orange!20}{\textbf{0.121}}
        & \colorbox{orange!20}{\textbf{23.88}} & \colorbox{orange!20}{\textbf{0.784}} & \colorbox{orange!20}{\textbf{0.123}} \\
        \bottomrule
    \end{tabular}}
    \caption{\textbf{Quantitative results of novel view synthesis on the synthetic dataset with 6 views.} The per-scene results sum to the average values provided.Each column is colored as: \colorbox{orange!20}{best} and \colorbox{topicblue!20}{second best}.}
    \label{app:tab:quantitative_per_scene_dyblurf1}
\end{table*}

\begin{table*}[h]
    \centering
    \footnotesize
    \setlength{\tabcolsep}{1.4pt}
    \renewcommand\arraystretch{1.2}
\resizebox{\linewidth}{!}{
    
    \begin{tabular}{@{}lcccccccccccccccccc@{}}
    \toprule
        \multirow{2}{*}{Methods} & \multicolumn{3}{c}{blurcozy2room} & \multicolumn{3}{c}{blurfactory} & \multicolumn{3}{c}{blurpool} & \multicolumn{3}{c}{blurtanabata} & \multicolumn{3}{c}{blurwine} & \multicolumn{3}{c}{Average} \\ 
        \cmidrule(r){2-4} \cmidrule(r){5-7} \cmidrule(r){8-10} \cmidrule(r){11-13} \cmidrule(r){14-16} \cmidrule(r){17-19}
        & \scriptsize PSNR$\uparrow$ & \scriptsize SSIM$\uparrow$ & \scriptsize LPIPS$\downarrow$
        & \scriptsize PSNR$\uparrow$ & \scriptsize SSIM$\uparrow$ & \scriptsize LPIPS$\downarrow$
        & \scriptsize PSNR$\uparrow$ & \scriptsize SSIM$\uparrow$ & \scriptsize LPIPS$\downarrow$
        & \scriptsize PSNR$\uparrow$ & \scriptsize SSIM$\uparrow$ & \scriptsize LPIPS$\downarrow$
        & \scriptsize PSNR$\uparrow$ & \scriptsize SSIM$\uparrow$ & \scriptsize LPIPS$\downarrow$
        & \scriptsize PSNR$\uparrow$ & \scriptsize SSIM$\uparrow$ & \scriptsize LPIPS$\downarrow$ \\ 
        \midrule
        BAD-Gaussians~\cite{zhao2024badgs}
        & \colorbox{topicblue!20}{28.45} & \colorbox{topicblue!20}{0.801} & \colorbox{topicblue!20}{0.092}
        & \colorbox{topicblue!20}{24.38} & \colorbox{topicblue!20}{0.866} & \colorbox{topicblue!20}{0.080}
        & \colorbox{topicblue!20}{27.05} & \colorbox{topicblue!20}{0.745} & \colorbox{topicblue!20}{0.139}
        & \colorbox{topicblue!20}{23.34} & \colorbox{topicblue!20}{0.794} & \colorbox{topicblue!20}{0.119}
        & \colorbox{topicblue!20}{23.92} & \colorbox{topicblue!20}{0.800} & \colorbox{topicblue!20}{0.090}
        & \colorbox{topicblue!20}{25.43} & \colorbox{topicblue!20}{0.801} & \colorbox{topicblue!20}{0.104} \\

        Sparse-DeRF$^*$~\cite{sparse-derf-tpami2025}
        & 25.71 & 0.768 & 0.147
        & 23.69 & 0.833 & 0.095
        & 25.66 & 0.714 & 0.127
        & 20.66 & 0.737 & 0.193
        & 20.55 & 0.679 & 0.153
        & 23.25 & 0.746 & 0.143 \\

        Difix3D+~\cite{wu2025difix3d+}
        & 22.66 & 0.749 & 0.183
        & 19.54 & 0.538 & 0.317
        & 21.25 & 0.604 & 0.254
        & 17.51 & 0.512 & 0.318
        & 17.75 & 0.517 & 0.301
        & 19.74 & 0.584 & 0.275 \\

        GenFusion~\cite{wu2025genfusion}
        & 21.89 & 0.744 & 0.341
        & 19.96 & 0.589 & 0.474
        & 18.51 & 0.602 & 0.439
        & 18.42 & 0.521 & 0.525
        & 18.58 & 0.548 & 0.481
        & 19.47 & 0.601 & 0.452 \\
        \midrule
        \textbf{Ours}
        & \colorbox{orange!20}{\text{29.37}} & \colorbox{orange!20}{\text{0.899}} & \colorbox{orange!20}{\text{0.050}}
        & \colorbox{orange!20}{\text{25.46}} & \colorbox{orange!20}{\text{0.891}} & \colorbox{orange!20}{\text{0.070}}
        & \colorbox{orange!20}{\text{27.78}} & \colorbox{orange!20}{\text{0.789}} & \colorbox{orange!20}{\text{0.122}}
        & \colorbox{orange!20}{\text{24.64}} & \colorbox{orange!20}{\text{0.837}} & \colorbox{orange!20}{\text{0.090}}
        & \colorbox{orange!20}{\text{24.56}} & \colorbox{orange!20}{\text{0.841}} & \colorbox{orange!20}{\text{0.070}}
        & \colorbox{orange!20}{\text{26.36}} & \colorbox{orange!20}{\text{0.851}} & \colorbox{orange!20}{\text{0.080}} \\
        \bottomrule
    \end{tabular}}
    \caption{\textbf{Quantitative results of novel view synthesis on the synthetic dataset with 9 views.}  The per-scene results sum to the average values provided. Each column is colored as: \colorbox{orange!20}{best} and \colorbox{topicblue!20}{second best}.}
    \label{app:tab:quantitative_per_scene_dyblurf2}
\end{table*}

\begin{table*}[h]
    \centering
    \footnotesize
    \setlength{\tabcolsep}{1.4pt}
    \renewcommand\arraystretch{1.2}
\resizebox{\linewidth}{!}{
    
    \begin{tabular}{@{}lcccccccccccccccccc@{}}
    \toprule
        \multirow{2}{*}{Methods} & \multicolumn{3}{c}{blurball} & \multicolumn{3}{c}{blurbuick} & \multicolumn{3}{c}{blurcoffee} & \multicolumn{3}{c}{blurdecoration} & \multicolumn{3}{c}{blurheron} & \multicolumn{3}{c}{Average} \\ 
        \cmidrule(r){2-4} \cmidrule(r){5-7} \cmidrule(r){8-10} \cmidrule(r){11-13} \cmidrule(r){14-16} \cmidrule(r){17-19}
        & \scriptsize PSNR$\uparrow$ & \scriptsize SSIM$\uparrow$ & \scriptsize LPIPS$\downarrow$
        & \scriptsize PSNR$\uparrow$ & \scriptsize SSIM$\uparrow$ & \scriptsize LPIPS$\downarrow$
        & \scriptsize PSNR$\uparrow$ & \scriptsize SSIM$\uparrow$ & \scriptsize LPIPS$\downarrow$
        & \scriptsize PSNR$\uparrow$ & \scriptsize SSIM$\uparrow$ & \scriptsize LPIPS$\downarrow$
        & \scriptsize PSNR$\uparrow$ & \scriptsize SSIM$\uparrow$ & \scriptsize LPIPS$\downarrow$
        & \scriptsize PSNR$\uparrow$ & \scriptsize SSIM$\uparrow$ & \scriptsize LPIPS$\downarrow$ \\ 
        \midrule
        BAD-Gaussians~\cite{zhao2024badgs}
        & 18.59 & 0.447 & 0.401
        & 15.88 & 0.412 & \colorbox{topicblue!20}{0.368}
        & \colorbox{topicblue!20}{25.70} & \colorbox{topicblue!20}{0.847} & \colorbox{topicblue!20}{0.176}
        & \colorbox{topicblue!20}{16.75} & 0.435 & 0.390
        & \colorbox{topicblue!20}{15.96} & 0.336 & \colorbox{topicblue!20}{0.353}
        & \colorbox{topicblue!20}{18.58} & 0.495 & \colorbox{topicblue!20}{0.338} \\

        Sparse-DeRF$^*$~\cite{sparse-derf-tpami2025}
        & 18.36 & 0.374 & 0.422
        & \colorbox{topicblue!20}{15.96} & 0.398 & 0.371
        & 23.16 & 0.708 & 0.267
        & 16.13 & 0.477 & \colorbox{topicblue!20}{0.382}
        & 15.28 & 0.357 & 0.452
        & 17.78 & 0.463 & 0.379 \\

        Difix3D+~\cite{wu2025difix3d+}
        & \colorbox{orange!20}{20.98} & \colorbox{orange!20}{0.556} & \colorbox{topicblue!20}{0.363}
        & 15.23 & \colorbox{topicblue!20}{0.458} & 0.471
        & 22.48 & 0.797 & 0.233
        & 16.22 & \colorbox{topicblue!20}{0.526} & 0.523
        & 15.38 & 0.330 & 0.485
        & 18.06 & \colorbox{topicblue!20}{0.533} & 0.415 \\

        GenFusion~\cite{wu2025genfusion}
        & 13.49 & 0.276 & 0.642
        & 13.02 & 0.375 & 0.543
        & 13.79 & 0.268 & 0.610
        & 12.65 & 0.438 & 0.573
        & 14.19 & \colorbox{topicblue!20}{0.377} & 0.575
        & 13.43 & 0.347 & 0.589 \\
        \midrule
        \textbf{Ours}
        & \colorbox{topicblue!20}{19.57} & \colorbox{topicblue!20}{0.525} & \colorbox{orange!20}{\text{0.358}}
        & \colorbox{orange!20}{\text{16.23}} & \colorbox{orange!20}{\text{0.493}} & \colorbox{orange!20}{\text{0.317}}
        & \colorbox{orange!20}{\text{26.95}} & \colorbox{orange!20}{\text{0.868}} & \colorbox{orange!20}{\text{0.152}}
        & \colorbox{orange!20}{\text{17.82}} & \colorbox{orange!20}{\text{0.535}} & \colorbox{orange!20}{0.378}
        & \colorbox{orange!20}{\text{17.32}} & \colorbox{orange!20}{\text{0.457}} & \colorbox{orange!20}{\text{0.281}}
        & \colorbox{orange!20}{\text{19.58}} & \colorbox{orange!20}{\text{0.576}} & \colorbox{orange!20}{\text{0.297}} \\
        \bottomrule
    \end{tabular}}
    \caption{\textbf{Quantitative results of novel view synthesis on the real-scene dataset with 3 views.} The per-scene results sum to the average values provided. Each column is colored as: \colorbox{orange!20}{best} and \colorbox{topicblue!20}{second best}.}
    \label{app:tab:3views_real}
\end{table*}

\begin{table*}[h]
    \centering
    \footnotesize
    \setlength{\tabcolsep}{1.4pt}
    \renewcommand\arraystretch{1.2}
\resizebox{\linewidth}{!}{
    
    \begin{tabular}{@{}lcccccccccccccccccc@{}}
    \toprule
        \multirow{2}{*}{Methods} & \multicolumn{3}{c}{blurball} & \multicolumn{3}{c}{blurbuick} & \multicolumn{3}{c}{blurcoffee} & \multicolumn{3}{c}{blurdecoration} & \multicolumn{3}{c}{blurheron} & \multicolumn{3}{c}{Average} \\ 
        \cmidrule(r){2-4} \cmidrule(r){5-7} \cmidrule(r){8-10} \cmidrule(r){11-13} \cmidrule(r){14-16} \cmidrule(r){17-19}
        & \scriptsize PSNR$\uparrow$ & \scriptsize SSIM$\uparrow$ & \scriptsize LPIPS$\downarrow$
        & \scriptsize PSNR$\uparrow$ & \scriptsize SSIM$\uparrow$ & \scriptsize LPIPS$\downarrow$
        & \scriptsize PSNR$\uparrow$ & \scriptsize SSIM$\uparrow$ & \scriptsize LPIPS$\downarrow$
        & \scriptsize PSNR$\uparrow$ & \scriptsize SSIM$\uparrow$ & \scriptsize LPIPS$\downarrow$
        & \scriptsize PSNR$\uparrow$ & \scriptsize SSIM$\uparrow$ & \scriptsize LPIPS$\downarrow$
        & \scriptsize PSNR$\uparrow$ & \scriptsize SSIM$\uparrow$ & \scriptsize LPIPS$\downarrow$ \\ 
        \midrule
        BAD-Gaussians~\cite{zhao2024badgs}
        & 20.94 & 0.562 & \colorbox{topicblue!20}{0.311}
        & \colorbox{topicblue!20}{18.62} & \colorbox{topicblue!20}{0.569} & \colorbox{topicblue!20}{0.271}
        & \colorbox{topicblue!20}{27.24} & \colorbox{topicblue!20}{0.872} & \colorbox{topicblue!20}{0.143}
        & 18.43 & 0.581 & 0.299
        & \colorbox{topicblue!20}{17.84} & 0.514 & 0.264
        & \colorbox{topicblue!20}{20.61} & \colorbox{topicblue!20}{0.620} & \colorbox{topicblue!20}{0.258} \\

        Sparse-DeRF$^*$~\cite{sparse-derf-tpami2025}
        & 20.62 & \colorbox{topicblue!20}{0.523} & 0.317
        & 18.16 & 0.525 & 0.291
        & 24.32 & 0.768 & 0.274
        & \colorbox{topicblue!20}{19.16} & \colorbox{topicblue!20}{0.626} & \colorbox{topicblue!20}{0.277}
        & 17.75 & \colorbox{topicblue!20}{0.549} & \colorbox{topicblue!20}{0.247}
        & 20.01 & 0.598 & 0.281 \\

        Difix3D+~\cite{wu2025difix3d+}
        & \colorbox{topicblue!20}{21.12} & 0.565 & 0.360
        & 17.20 & 0.510 & 0.420
        & 24.90 & 0.844 & 0.205
        & 17.48 & 0.553 & 0.516
        & 16.82 & 0.397 & 0.446
        & 19.50 & 0.574 & 0.389 \\

        GenFusion~\cite{wu2025genfusion}
        & 18.26 & 0.488 & 0.583
        & 17.66 & 0.566 & 0.431
        & 15.95 & 0.385 & 0.563
        & 12.56 & 0.489 & 0.535
        & 15.32 & 0.438 & 0.554
        & 15.95 & 0.473 & 0.533 \\
        \midrule
        \textbf{Ours}
        & \colorbox{orange!20}{\text{22.31}} & \colorbox{orange!20}{\text{0.634}} & \colorbox{orange!20}{\text{0.259}}
        & \colorbox{orange!20}{\text{19.98}} & \colorbox{orange!20}{\text{0.662}} & \colorbox{orange!20}{\text{0.189}}
        & \colorbox{orange!20}{\text{29.77}} & \colorbox{orange!20}{\text{0.919}} & \colorbox{orange!20}{\text{0.090}}
        & \colorbox{orange!20}{\text{19.26}} & \colorbox{orange!20}{\text{0.631}} & \colorbox{orange!20}{\text{0.252}}
        & \colorbox{orange!20}{\text{18.69}} & \colorbox{orange!20}{\text{0.560}} & \colorbox{orange!20}{\text{0.218}}
        & \colorbox{orange!20}{\text{22.00}} & \colorbox{orange!20}{\text{0.681}} & \colorbox{orange!20}{\text{0.202}} \\
        \bottomrule
    \end{tabular}}
    \caption{\textbf{Quantitative results of novel view synthesis on the real-scene dataset with 6 views.} The per-scene results sum to the average values provided. Each column is colored as: \colorbox{orange!20}{best} and \colorbox{topicblue!20}{second best}.}
    \label{app:tab:quantitative_per_scene_dyblurf3}
\end{table*}

\begin{table*}[h]
    \centering
    \footnotesize
    \setlength{\tabcolsep}{1.4pt}
    \renewcommand\arraystretch{1.2}
\resizebox{\linewidth}{!}{
    
    \begin{tabular}{@{}lcccccccccccccccccc@{}}
    \toprule
        \multirow{2}{*}{Methods} & \multicolumn{3}{c}{blurball} & \multicolumn{3}{c}{blurbuick} & \multicolumn{3}{c}{blurcoffee} & \multicolumn{3}{c}{blurdecoration} & \multicolumn{3}{c}{blurheron} & \multicolumn{3}{c}{Average} \\ 
        \cmidrule(r){2-4} \cmidrule(r){5-7} \cmidrule(r){8-10} \cmidrule(r){11-13} \cmidrule(r){14-16} \cmidrule(r){17-19}
        & \scriptsize PSNR$\uparrow$ & \scriptsize SSIM$\uparrow$ & \scriptsize LPIPS$\downarrow$
        & \scriptsize PSNR$\uparrow$ & \scriptsize SSIM$\uparrow$ & \scriptsize LPIPS$\downarrow$
        & \scriptsize PSNR$\uparrow$ & \scriptsize SSIM$\uparrow$ & \scriptsize LPIPS$\downarrow$
        & \scriptsize PSNR$\uparrow$ & \scriptsize SSIM$\uparrow$ & \scriptsize LPIPS$\downarrow$
        & \scriptsize PSNR$\uparrow$ & \scriptsize SSIM$\uparrow$ & \scriptsize LPIPS$\downarrow$
        & \scriptsize PSNR$\uparrow$ & \scriptsize SSIM$\uparrow$ & \scriptsize LPIPS$\downarrow$ \\ 
        \midrule
        BAD-Gaussians~\cite{zhao2024badgs}
        & \colorbox{topicblue!20}{22.65} & \colorbox{topicblue!20}{0.648} & \colorbox{topicblue!20}{0.257}
        & \colorbox{topicblue!20}{19.75} & \colorbox{topicblue!20}{0.567} & \colorbox{topicblue!20}{0.241}
        & \colorbox{topicblue!20}{28.38} & \colorbox{topicblue!20}{0.888} & \colorbox{topicblue!20}{0.133}
        & 20.12 & 0.666 & 0.242
        & \colorbox{topicblue!20}{19.08} & \colorbox{topicblue!20}{0.529} & \colorbox{topicblue!20}{0.209}
        & \colorbox{topicblue!20}{22.00} & \colorbox{topicblue!20}{0.660} & \colorbox{topicblue!20}{0.216} \\

        Sparse-DeRF$^*$~\cite{sparse-derf-tpami2025}
        & 21.79 & 0.645 & 0.297
        & 18.72 & 0.521 & 0.290
        & 26.37 & 0.748 & 0.166
        & \colorbox{topicblue!20}{20.39} & \colorbox{topicblue!20}{0.692} & \colorbox{topicblue!20}{0.227}
        & 18.55 & 0.483 & 0.231
        & 21.16 & 0.618 & 0.242 \\

        Difix3D+~\cite{wu2025difix3d+}
        & 20.57 & 0.539 & 0.328
        & 17.97 & 0.548 & 0.402
        & 24.92 & 0.848 & 0.201
        & 18.39 & 0.578 & 0.462
        & 17.08 & 0.410 & 0.446
        & 19.79 & 0.585 & 0.368 \\

        GenFusion~\cite{wu2025genfusion}
        & 16.92 & 0.436 & 0.610
        & 17.57 & 0.551 & 0.455
        & 17.08 & 0.401 & 0.564
        & 12.97 & 0.490 & 0.548
        & 16.41 & 0.486 & 0.524
        & 16.19 & 0.473 & 0.540 \\
        \midrule
        \textbf{Ours}
        & \colorbox{orange!20}{\text{24.90}} & \colorbox{orange!20}{\text{0.743}} & \colorbox{orange!20}{\text{0.220}}
        & \colorbox{orange!20}{\text{21.71}} & \colorbox{orange!20}{\text{0.758}} & \colorbox{orange!20}{\text{0.142}}
        & \colorbox{orange!20}{\text{30.43}} & \colorbox{orange!20}{\text{0.921}} & \colorbox{orange!20}{\text{0.086}}
        & \colorbox{orange!20}{\text{20.64}} & \colorbox{orange!20}{\text{0.705}} & \colorbox{orange!20}{\text{0.194}}
        & \colorbox{orange!20}{\text{19.52}} & \colorbox{orange!20}{\text{0.612}} & \colorbox{orange!20}{\text{0.193}}
        & \colorbox{orange!20}{\text{23.44}} & \colorbox{orange!20}{\text{0.748}} & \colorbox{orange!20}{\text{0.167}} \\
        \bottomrule
    \end{tabular}}
    \caption{\textbf{Quantitative results of novel view synthesis on the real-scene dataset with 9 views.} The per-scene results sum to the average values provided. Each column is colored as: \colorbox{orange!20}{best} and \colorbox{topicblue!20}{second best}.}
    \label{app:tab:quantitative_per_scene_dyblurf4}
\end{table*}

\begin{table*}[h]
    \centering
    \footnotesize
    \setlength{\tabcolsep}{1.4pt}
    \renewcommand\arraystretch{1.2}
\resizebox{\linewidth}{!}{
    
    \begin{tabular}{@{}lcccccccccccccccccc@{}}
    \toprule
        \multirow{2}{*}{Methods} & \multicolumn{3}{c}{0001} & \multicolumn{3}{c}{0002} & \multicolumn{3}{c}{0003} & \multicolumn{3}{c}{0004} & \multicolumn{3}{c}{0005} & \multicolumn{3}{c}{Average} \\ 
        \cmidrule(r){2-4} \cmidrule(r){5-7} \cmidrule(r){8-10} \cmidrule(r){11-13} \cmidrule(r){14-16} \cmidrule(r){17-19}
        & \scriptsize PSNR$\uparrow$ & \scriptsize SSIM$\uparrow$ & \scriptsize LPIPS$\downarrow$
        & \scriptsize PSNR$\uparrow$ & \scriptsize SSIM$\uparrow$ & \scriptsize LPIPS$\downarrow$
        & \scriptsize PSNR$\uparrow$ & \scriptsize SSIM$\uparrow$ & \scriptsize LPIPS$\downarrow$
        & \scriptsize PSNR$\uparrow$ & \scriptsize SSIM$\uparrow$ & \scriptsize LPIPS$\downarrow$
        & \scriptsize PSNR$\uparrow$ & \scriptsize SSIM$\uparrow$ & \scriptsize LPIPS$\downarrow$
        & \scriptsize PSNR$\uparrow$ & \scriptsize SSIM$\uparrow$ & \scriptsize LPIPS$\downarrow$ \\ 
        \midrule
        BAD-Gaussians~\cite{zhao2024badgs}
        & 15.60 & 0.513 & 0.431
        & \colorbox{topicblue!20}{15.08} & 0.484 & 0.536
        & 14.21 & 0.389 & 0.517
        & 12.55 & 0.517 & 0.501
        & \colorbox{topicblue!20}{17.95} & \colorbox{topicblue!20}{0.602} & \colorbox{topicblue!20}{0.346}
        & \colorbox{topicblue!20}{15.08} & 0.501 & 0.466 \\

        Sparse-DeRF$^*$~\cite{sparse-derf-tpami2025}
        & \colorbox{topicblue!20}{15.68} & \colorbox{topicblue!20}{0.565} & \colorbox{topicblue!20}{0.412}
        & 15.01 & 0.425 & 0.560
        & 13.32 & 0.328 & 0.511
        & 12.79 & 0.508 & 0.497
        & 17.39 & {0.589} & 0.527
        & 14.84 & 0.483 & 0.501 \\

        Difix3D+~\cite{wu2025difix3d+}
        & 13.90 & 0.475 & 0.430
        & 14.71 & 0.439 & \colorbox{orange!20}{0.415}
        & \colorbox{topicblue!20}{14.66} & \colorbox{topicblue!20}{0.422} & \colorbox{orange!20}{0.423}
        & \colorbox{topicblue!20}{12.88} & 0.528 & \colorbox{topicblue!20}{0.396}
        & 15.66 & 0.596 & 0.358
        & 14.36 & 0.492 & \colorbox{topicblue!20}{0.404} \\

        GenFusion~\cite{wu2025genfusion}
        & 12.11 & 0.452 & 0.563
        & 13.24 & \colorbox{topicblue!20}{0.521} & 0.582
        & 13.68 & 0.421 & 0.585
        & 12.34 & \colorbox{topicblue!20}{0.551} & 0.459
        & 12.73 & 0.568 & 0.519
        & 12.82 & \colorbox{topicblue!20}{0.503} & 0.542 \\
        \midrule
        \textbf{Ours}
        & \colorbox{orange!20}{18.26} & \colorbox{orange!20}{\text{0.657}} & \colorbox{orange!20}{\text{0.322}}
        & \colorbox{orange!20}{16.80} & \colorbox{orange!20}{\text{0.607}} & \colorbox{topicblue!20}{\text{0.434}}
        & \colorbox{orange!20}{15.67} & \colorbox{orange!20}{\text{0.535}} & \colorbox{topicblue!20}{0.489}
        & \colorbox{orange!20}{14.73} & \colorbox{orange!20}{\text{0.597}} & \colorbox{orange!20}{\text{0.378}}
        & \colorbox{orange!20}{21.95} & \colorbox{orange!20}{\text{0.801}} & \colorbox{orange!20}{\text{0.217}}
        & \colorbox{orange!20}{\text{17.48}} & \colorbox{orange!20}{\text{0.639}} & \colorbox{orange!20}{\text{0.368}} \\
        \bottomrule
    \end{tabular}}
    \caption{\textbf{Quantitative results of novel view synthesis on the DL3DV-Blur dataset with 3 views.} The per-scene results sum to the average values provided. Each column is colored as: \colorbox{orange!20}{best} and \colorbox{topicblue!20}{second best}.}
    \label{app:tab:quantitative_per_scene_dyblurf5}
\end{table*}

\begin{table*}[h]
    \centering
    \footnotesize
    \setlength{\tabcolsep}{1.4pt}
    \renewcommand\arraystretch{1.2}
\resizebox{\linewidth}{!}{
    
    \begin{tabular}{@{}lcccccccccccccccccc@{}}
    \toprule
        \multirow{2}{*}{Methods} & \multicolumn{3}{c}{0001} & \multicolumn{3}{c}{0002} & \multicolumn{3}{c}{0003} & \multicolumn{3}{c}{0004} & \multicolumn{3}{c}{0005} & \multicolumn{3}{c}{Average} \\ 
        \cmidrule(r){2-4} \cmidrule(r){5-7} \cmidrule(r){8-10} \cmidrule(r){11-13} \cmidrule(r){14-16} \cmidrule(r){17-19}
        & \scriptsize PSNR$\uparrow$ & \scriptsize SSIM$\uparrow$ & \scriptsize LPIPS$\downarrow$
        & \scriptsize PSNR$\uparrow$ & \scriptsize SSIM$\uparrow$ & \scriptsize LPIPS$\downarrow$
        & \scriptsize PSNR$\uparrow$ & \scriptsize SSIM$\uparrow$ & \scriptsize LPIPS$\downarrow$
        & \scriptsize PSNR$\uparrow$ & \scriptsize SSIM$\uparrow$ & \scriptsize LPIPS$\downarrow$
        & \scriptsize PSNR$\uparrow$ & \scriptsize SSIM$\uparrow$ & \scriptsize LPIPS$\downarrow$
        & \scriptsize PSNR$\uparrow$ & \scriptsize SSIM$\uparrow$ & \scriptsize LPIPS$\downarrow$ \\ 
        \midrule
        BAD-Gaussians~\cite{zhao2024badgs}
        & \colorbox{topicblue!20}{19.49} & \colorbox{topicblue!20}{0.643} & \colorbox{topicblue!20}{0.289}
        & 18.35 & 0.586 & \colorbox{topicblue!20}{0.359}
        & 16.07 & 0.471 & 0.432
        & 16.77 & \colorbox{topicblue!20}{0.647} & 0.309
        & \colorbox{topicblue!20}{22.43} & \colorbox{topicblue!20}{0.826} & \colorbox{topicblue!20}{0.176}
        & \colorbox{topicblue!20}{18.62} & \colorbox{topicblue!20}{0.635} & \colorbox{topicblue!20}{0.313} \\

        Sparse-DeRF$^*$~\cite{sparse-derf-tpami2025}
        & 19.35 & 0.618 & 0.301
        & \colorbox{topicblue!20}{18.61} & 0.611 & 0.364
        & \colorbox{topicblue!20}{16.11} & \colorbox{topicblue!20}{0.508} & 0.445
        & \colorbox{topicblue!20}{17.11} & 0.625 & \colorbox{topicblue!20}{0.289}
        & 21.90 & 0.729 & 0.227
        & 18.61 & 0.618 & 0.325 \\

        Difix3D+~\cite{wu2025difix3d+}
        & 16.31 & 0.552 & 0.638
        & 16.27 & 0.519 & 0.363
        & 14.69 & 0.421 & \colorbox{topicblue!20}{0.413}
        & 14.46 & 0.604 & 0.304
        & 15.93 & 0.597 & 0.362
        & 15.53 & 0.538 & 0.416 \\

        GenFusion~\cite{wu2025genfusion}
        & 15.83 & 0.605 & 0.486
        & 17.36 & \colorbox{topicblue!20}{0.621} & 0.506
        & 14.75 & 0.463 & 0.536
        & 16.13 & 0.616 & 0.448
        & 18.04 & 0.740 & 0.415
        & 16.42 & 0.609 & 0.478 \\
        \midrule
        \textbf{Ours}
        & \colorbox{orange!20}{\text{21.89}} & \colorbox{orange!20}{\text{0.704}} & \colorbox{orange!20}{\text{0.234}}
        & \colorbox{orange!20}{\text{19.06}} & \colorbox{orange!20}{\text{0.636}} & \colorbox{orange!20}{\text{0.314}}
        & \colorbox{orange!20}{\text{17.32}} & \colorbox{orange!20}{\text{0.512}} & \colorbox{orange!20}{\text{0.393}}
        & \colorbox{orange!20}{\text{17.46}} & \colorbox{orange!20}{\text{0.695}} & \colorbox{orange!20}{\text{0.254}}
        & \colorbox{orange!20}{\textbf{23.77}} & \colorbox{orange!20}{\text{0.847}} & \colorbox{orange!20}{\text{0.144}}
        & \colorbox{orange!20}{\text{19.90}} & \colorbox{orange!20}{\text{0.679}} & \colorbox{orange!20}{\text{0.268}} \\
        \bottomrule
    \end{tabular}}
    \caption{\textbf{Quantitative results of novel view synthesis on the DL3DV-Blur dataset with 6 views.} The per-scene results sum to the average values provided. Each column is colored as: \colorbox{orange!20}{best} and \colorbox{topicblue!20}{second best}.}
    \label{app:tab:dl3dv_blur_6views}
\end{table*}

\begin{table*}[h]
    \centering
    \footnotesize
    \setlength{\tabcolsep}{1.4pt}
    \renewcommand\arraystretch{1.2}
\resizebox{\linewidth}{!}{
    
    \begin{tabular}{@{}lcccccccccccccccccc@{}}
    \toprule
        \multirow{2}{*}{Methods} & \multicolumn{3}{c}{0001} & \multicolumn{3}{c}{0002} & \multicolumn{3}{c}{0003} & \multicolumn{3}{c}{0004} & \multicolumn{3}{c}{0005} & \multicolumn{3}{c}{Average} \\ 
        \cmidrule(r){2-4} \cmidrule(r){5-7} \cmidrule(r){8-10} \cmidrule(r){11-13} \cmidrule(r){14-16} \cmidrule(r){17-19}
        & \scriptsize PSNR$\uparrow$ & \scriptsize SSIM$\uparrow$ & \scriptsize LPIPS$\downarrow$
        & \scriptsize PSNR$\uparrow$ & \scriptsize SSIM$\uparrow$ & \scriptsize LPIPS$\downarrow$
        & \scriptsize PSNR$\uparrow$ & \scriptsize SSIM$\uparrow$ & \scriptsize LPIPS$\downarrow$
        & \scriptsize PSNR$\uparrow$ & \scriptsize SSIM$\uparrow$ & \scriptsize LPIPS$\downarrow$
        & \scriptsize PSNR$\uparrow$ & \scriptsize SSIM$\uparrow$ & \scriptsize LPIPS$\downarrow$
        & \scriptsize PSNR$\uparrow$ & \scriptsize SSIM$\uparrow$ & \scriptsize LPIPS$\downarrow$ \\ 
        \midrule
        BAD-Gaussians~\cite{zhao2024badgs}
        & 22.56 & 0.733 & \colorbox{topicblue!20}{0.213}
        & 18.81 & 0.617 & 0.321
        & \colorbox{topicblue!20}{17.68} & \colorbox{topicblue!20}{0.525} & 0.369
        & 17.24 & 0.633 & \colorbox{topicblue!20}{0.270}
        & \colorbox{topicblue!20}{22.80} & \colorbox{topicblue!20}{0.845} & \colorbox{topicblue!20}{0.161}
        & \colorbox{topicblue!20}{19.82} & 0.671 & \colorbox{topicblue!20}{0.267} \\

        Sparse-DeRF$^*$~\cite{sparse-derf-tpami2025}
        & \colorbox{topicblue!20}{22.76} & \colorbox{topicblue!20}{0.745} & 0.235
        & \colorbox{topicblue!20}{19.21} & \colorbox{topicblue!20}{0.693} & \colorbox{topicblue!20}{0.291}
        & 17.25 & 0.508 & 0.386
        & 17.39 & 0.612 & 0.278
        & 21.95 & 0.816 & 0.196
        & 19.72 & \colorbox{topicblue!20}{0.674} & 0.277 \\

        Difix3D+~\cite{wu2025difix3d+}
        & 16.95 & 0.628 & 0.310
        & 14.86 & 0.553 & 0.369
        & 17.30 & 0.505 & \colorbox{topicblue!20}{0.351}
        & 12.99 & 0.577 & 0.398
        & 18.69 & 0.706 & 0.251
        & 16.16 & 0.594 & 0.336 \\

        GenFusion~\cite{wu2025genfusion}
        & 18.87 & 0.656 & 0.437
        & 18.11 & 0.654 & 0.585
        & 15.70 & 0.476 & 0.520
        & \colorbox{topicblue!20}{17.83} & \colorbox{topicblue!20}{0.682} & 0.416
        & 18.67 & 0.765 & 0.391
        & 17.84 & 0.647 & 0.470 \\
        \midrule
        \textbf{Ours}
        & \colorbox{orange!20}{\text{24.32}} & \colorbox{orange!20}{\text{0.785}} & \colorbox{orange!20}{\text{0.177}}
        & \colorbox{orange!20}{\text{20.99}} & \colorbox{orange!20}{\text{0.701}} & \colorbox{orange!20}{\text{0.266}}
        & \colorbox{orange!20}{\text{18.81}} & \colorbox{orange!20}{\text{0.582}} & \colorbox{orange!20}{\text{0.332}}
        & \colorbox{orange!20}{\text{19.60}} & \colorbox{orange!20}{\text{0.760}} & \colorbox{orange!20}{\text{0.211}}
        & \colorbox{orange!20}{\text{24.51}} & \colorbox{orange!20}{\text{0.871}} & \colorbox{orange!20}{\text{0.129}}
        & \colorbox{orange!20}{\text{21.65}} & \colorbox{orange!20}{\text{0.740}} & \colorbox{orange!20}{\text{0.223}} \\
        \bottomrule
    \end{tabular}}
    \caption{\textbf{Quantitative results of novel view synthesis on the DL3DV-Blur dataset with 9 views.} The per-scene results sum to the average values provided. Each column is colored as: \colorbox{orange!20}{best} and \colorbox{topicblue!20}{second best}.}
    \label{app:tab:dl3dv_blur_9views}
\end{table*}

%% file: Tables/table_training_indices.tex
\begin{table}[t!]
\centering
\caption{Image indices in training settings.}
\label{tab:app:traing indices}
\begin{tabular}{lccc}
\toprule
\textbf{View selection} & 3views & 6views & 9views \\
\midrule
{\textbf{DL3DV-BLUR}} & 5,15,25 & 2,5,10,15,17,25 &1,2,5,10,15,17,22,25 \\
{\textbf{Deblur-NeRF}} & 5,15,25 & 2,5,10,15,17,25 &1,2,5,10,15,17,22,25 \\

\bottomrule
\end{tabular}
\vspace{-10pt}
\end{table}